\RequirePackage{lineno}
\documentclass[preprint,12pt,authoryear]{elsarticle}

\usepackage{amssymb}

\usepackage{amsmath}

\usepackage{booktabs}
\usepackage{makecell}
\usepackage{multirow}
\usepackage{graphicx}

\usepackage{hyperref}

\usepackage{float}

\usepackage{lineno}

\begin{document}

\begin{frontmatter}



\title{A Spatially Masked Adaptive Gated Network for multimodal post‑flood water extent mapping using SAR and incomplete multispectral data}

\author{Hyunho Lee\fnref{asu}}
\author{Wenwen Li\corref{cor}\fnref{asu}}
\ead{wenwen@asu.edu}
\affiliation[asu]{organization={School of Geographical Sciences and Urban Planning, Arizona State University},
           city={Tempe},
           postcode={85287-5302}, 
           state={AZ},
           country={USA}}
\cortext[cor]{Corresponding author.}




\begin{abstract}
Mapping water extent during a flood event is essential for effective disaster management throughout all phases: mitigation, preparedness, response, and recovery. In particular, during the response stage, when timely and accurate information is important, Synthetic Aperture Radar (SAR) data are primarily employed to produce water extent maps. This is because SAR sensors can observe through cloud cover and operate both day and night, whereas Multispectral Imaging (MSI) data, despite providing higher mapping accuracy, are only available under cloud-free and daytime conditions. Recently, leveraging the complementary characteristics of SAR and MSI data through a multimodal approach has emerged as a promising strategy for advancing water extent mapping using deep learning models. This approach is particularly beneficial when timely post-flood observations, acquired during or shortly after the flood peak, are limited, as it enables the use of all available imagery for more accurate post-flood water extent mapping. However, the adaptive integration of partially available MSI data into the SAR-based post-flood water extent mapping process remains underexplored. To bridge this research gap, we propose the Spatially Masked Adaptive Gated Network (SMAGNet), a multimodal deep learning model that utilizes SAR data as the primary input for post-flood water extent mapping and integrates complementary MSI data through feature fusion. In experiments on the C2S-MS Floods dataset, SMAGNet consistently outperformed other multimodal deep learning models in prediction performance across varying levels of MSI data availability. Specifically, SMAGNet achieved the highest IoU score of 86.47\% using SAR and MSI data and maintained the highest performance with an IoU score of 79.53\% even when MSI data were entirely missing. Furthermore, we found that even when MSI data were completely missing, the performance of SMAGNet remained statistically comparable to that of a U-Net model trained solely on SAR data. These findings indicate that SMAGNet enhances the model robustness to missing data as well as the applicability of multimodal deep learning in real-world flood management scenarios. The source code is available at https://github.com/ASUcicilab/SMAGNet.
\end{abstract}



\begin{keyword}
Multimodal deep learning \sep Flood mapping \sep Missing data \sep Spatial mask \sep Feature fusion 



\end{keyword}

\end{frontmatter}


\section{Introduction}
\label{Section1}

Climate change is projected to increase the frequency and intensity of extreme precipitation events, which are likely to exacerbate the severity of flooding \citep{najibi2018recent, shu2023integrating, tabari2020climate}. In light of these projections, flood maps depicting inundation extent, depth, vulnerability, and risk \citep{bentivoglio2022deep, cova1999gis} are becoming increasingly crucial for effective spatial decision-making across all phases of flood management: mitigation, preparedness, response, and recovery \citep{ajmar2017response}. In particular, during the response phase, flood extent mapping with satellite data is an essential task that provides timely information on flood-affected areas to decision-makers \citep{wania2021increasing}.

Specifically, a flood extent map refers to a type of map that delineates the area affected by a flood \citep{hashemi2021flood, wang2002mapping}. In flood extent maps, flooded areas are typically identified  by subtracting the permanent or pre-flood water extent from the post-flood water extent \citep{ajmar2017response, he2023cross, saleh2024dam}. Post-flood water extent mapping generally utilizes satellite imagery acquired during or shortly after the flood peak to reflect the maximum flood extent \citep{huang2018near, misra2025mapping, samela2022satellite, vanama2021change}. In this process, post-flood water extent mapping is critical, as it not only provides spatial information on the extent of water bodies after a flood event, which is essential for water resource management \citep{risling2024comparison}, but also plays a key role in producing accurate flood extent maps from the perspective of disaster management. For brevity, this study refers to the mapping of water extent using post-flood satellite imagery acquired during or shortly after the flood peak as post-flood water mapping.

Synthetic Aperture Radar (SAR) and Multispectral Imaging (MSI) data are the primary satellite data sources utilized for post-flood water mapping, with each providing complementary capabilities \citep{konapala2021exploring}. In particular, SAR data are effectively leveraged during flood response stages to produce timely water extent maps due to their ability to provide observations of the Earth's surface in all weather conditions and at any time of day \citep{ajmar2017response, boccardo2015remote, chaouch2012synergetic, uddin2019operational}. This capability is enabled by SAR sensors detecting scattered energy from emitted microwave pulses. The amount of scattered energy is primarily determined by surface roughness \citep{grimaldi2020flood}. Rough land surfaces scatter energy in multiple directions, including back toward the sensor, causing high backscatter. In contrast, open water surfaces reflect radar signals away from the sensor, resulting in low backscatter. However, using SAR data alone in post-flood water mapping faces some limitations, including speckle noise, difficulty distinguishing man-made flat surfaces (e.g., roads, airport runways) from open water, and double-bounce backscattering from buildings and flooded vegetation \citep{amitrano2024flood, grimaldi2020flood}. On the other hand, although MSI data have limitations in observational availability caused by cloud cover, they provide water-sensitive spectral bands such as Near Infrared (NIR) and Shortwave Infrared (SWIR) under cloud-free conditions, which significantly enhance the accuracy of post-flood water mapping \citep{konapala2021exploring}. In addition, due to ease of visual interpretation, MSI data are predominantly utilized to assess flood-induced damage to infrastructure, such as buildings and roads \citep{ajmar2017response, boccardo2015remote}.

Leveraging the complementary characteristics of both SAR and MSI data through a multimodal approach is a promising research direction for advancing post-flood water mapping research using deep learning models \citep{bentivoglio2022deep, li2024geoai, rolf2024mission}. In this context, modality refers to a distinct type of data acquired from a single sensor in the observation of a phenomenon or system \citep{li2025multi, ramachandram2017deep}. Deep learning has particular strengths in recognizing patterns from multimodal satellite data by learning complex relationships between modalities through end-to-end optimization. Specifically, in contrast to rule-based methods, which rely on predefined thresholds and rules, and traditional machine learning, which requires feature engineering, deep learning reduces the dependence on heuristic decisions in the modeling process \citep{amitrano2024flood, li2022water, wieland2019modular}, when utilizing multimodal satellite data.

Recently, considerable research has been conducted on deep learning using multimodal satellite data \citep{hosseinpour2022cmgfnet, li2022mcanet, liu2024review, ma2024multilevel, mena2024common, sun2021deep, yu2024cmfpnet, zhao2022multi}, including applications in post-flood water mapping and flood mapping \citep{drakonakis2022ombrianet, he2023cross, konapala2021exploring, sanderson2023optimal}. Notably, the previous study \citep{konapala2021exploring} has shown that the integration of MSI data with SAR data can improve the accuracy of post-flood water mapping. However, in real-world scenarios of a multimodal deep learning for SAR-based post-flood water mapping, acquiring fully available MSI data as model inputs that capture the same location within a short time interval as SAR data is not always feasible. This is because MSI data utilized as supplementary input for SAR-based post-flood water mapping often contains missing data pixels due to factors such as limited temporal resolution of satellite sensors, coregistration process between SAR and MSI data, sensor swath constraints, errors during transmission, and potential sensor malfunctions. Despite this limitation, most deep learning studies utilizing multimodal satellite data either assume that all data modalities are fully available \citep{hosseinpour2022cmgfnet, liu2024review, mena2024common} or consider availability at the modality-level \citep{adriano2021learning, kampffmeyer2018urban, li2021dynamic, liu2024multimodal, wei2023msh}, without addressing pixel-level availability issues. Consequently, the adaptive integration of partially available MSI data into the SAR-based post-flood water mapping process through multimodal deep learning remains underexplored. 

To bridge this research gap, we propose the Spatially Masked Adaptive Gated Network (SMAGNet), a novel multimodal deep learning model designed to improve the accuracy of SAR-based post-flood water mapping during the flood response phase by integrating MSI data to leverage their complementary features and simultaneously addressing issues of missing data. Our experiments demonstrate that SMAGNet not only outperforms other multimodal deep learning models but also maintains robustness in the presence of missing data pixels in MSI data. The main contributions of this study are:

\begin{enumerate}
    \renewcommand{\labelenumi}{\arabic{enumi})}  
    \item We introduce a novel Spatially Masked Adaptive Gated Network (SMAGNet) to adaptively integrate partially available MSI data into the SAR-based post-flood water mapping process based on multimodal deep learning.
    \item We demonstrate the superior performance of SMAGNet in post-flood water mapping compared to other multimodal deep learning models through comprehensive experimental results.
    \item Furthermore, we found that even when MSI data were completely missing, the performance of SMAGNet remained statistically comparable to that of a U-Net model trained solely on SAR data. This finding indicates that our method enhances the model robustness to missing data pixels in MSI data and applicability of multimodal deep learning in real-world flood management scenarios.
\end{enumerate}

The structure of this paper is as follows: Section \ref{Section2} reviews relevant literature; Section \ref{Section3} details the architecture of the proposed model; Section \ref{Section4} outlines the experimental setup, and Section \ref{Section5} presents the results; Section \ref{Section6} provides a discussion, including a comparative analysis, robustness evaluations, an ablation study, and a generalizability study; Finally, Section \ref{Section7} summarizes the findings and suggests directions for future research.

\section{Literature Review}
\label{Section2}

\subsection{Multimodal Deep Learning with Geospatial Data}
\label{Section2.1}
Multimodal deep learning has been actively explored to enhance the accuracy of Earth observation and mapping tasks by integrating various types of geospatial data. Consequently, considerable research has been directed toward developing advanced deep learning architectures and fusion techniques to effectively combine multiple geospatial modalities, such as satellite imagery, digital elevation models (DEM), digital surface models (DSM), and LiDAR data \citep{huang2023deep, rolf2024mission}. In previous studies, three key aspects have primarily been considered when designing these multimodal deep learning models: (1) the selection of data sources, (2) the stages of fusion within the model, and (3) the fusion methods \citep{huang2023deep, kang2022cfnet, mena2024common}.

First, in terms of data source selection, particularly for post-flood water mapping and flood mapping, SAR and MSI data are the most commonly utilized sources \citep{bonafilia2020sen1floods11, Cloud_to_Street2022, drakonakis2022ombrianet, he2023cross, konapala2021exploring, montello2022mmflood, sanderson2023optimal, zhang2023new}. Table \ref{tab1}  presents publicly accessible datasets for multimodal deep learning, explicitly developed for post-flood water mapping using multiple geospatial data. These datasets are all designed for the semantic segmentation task and contain globally distributed data to enhance the generalizability of deep learning models by covering diverse vegetation types, climates, and regions. However, each dataset employs different labeling methods tailored to its specific purpose.

\begin{table}[t]
\centering
\caption{Summary of datasets for multimodal deep learning in post-flood water mapping. In timestamps, Pre means the pre-flood event phase, and Post indicates the post-flood event phase.}
\label{tab1}
\resizebox{\textwidth}{!}{%
\begin{tabular}{ccccccc}
\toprule 
\textbf{\begin{tabular}[c]{@{}c@{}}Dataset\\ (Reference)\end{tabular}} &
  \textbf{Modality} &
  \textbf{Timestamps} &
  \textbf{File format} &
  \textbf{\begin{tabular}[c]{@{}c@{}}\# of \\ flood event \end{tabular}} &
  \textbf{\begin{tabular}[c]{@{}c@{}}\# of pairs\\ (Image size)\end{tabular}} &
  \textbf{Labeling method} \\ \hline\hline  
\begin{tabular}[c]{@{}c@{}}Sen1Floods11 \\ \citep{bonafilia2020sen1floods11} \end{tabular} &
  SAR, MSI &
  Post &
  GeoTiff &
  \begin{tabular}[c]{@{}c@{}}11\end{tabular} &
  \begin{tabular}[c]{@{}c@{}}446\\ (512$\times$512)\end{tabular} &
  \begin{tabular}[c]{@{}c@{}}Manually annotated\\ pixel-level labels by \\ combining SAR and MSI\end{tabular} \\ \hline
\begin{tabular}[c]{@{}c@{}}C2S-MS Floods \\ \citep{Cloud_to_Street2022} \end{tabular} &
  SAR, MSI &
  Post &
  GeoTiff &
  \begin{tabular}[c]{@{}c@{}}18\end{tabular} &
  \begin{tabular}[c]{@{}c@{}}900\\ (512$\times$512)\end{tabular} &
  \begin{tabular}[c]{@{}c@{}}Manually annotated\\ pixel-level labels\\ separately (SAR,  MSI)\end{tabular} \\ \hline
\begin{tabular}[c]{@{}c@{}}MM-Flood \\ \citep{montello2022mmflood} \end{tabular} &
  \begin{tabular}[c]{@{}c@{}}SAR, DEM,\\ hydrography\\ map\end{tabular} &
  Post &
  GeoTiff &
  \begin{tabular}[c]{@{}c@{}}95\end{tabular} &
  \begin{tabular}[c]{@{}c@{}}1,748\\ (2,000$\times$2,000)\end{tabular} &
  \begin{tabular}[c]{@{}c@{}}Pixel-level labels from\\ EMS (Emergency Management\\ Service) polygons\end{tabular} \\ \hline
\begin{tabular}[c]{@{}c@{}}Ombria \\ \citep{drakonakis2022ombrianet} \end{tabular} &
  SAR, MSI &
  \begin{tabular}[c]{@{}c@{}}Pre (SAR,\\ MSI), Post\\ (SAR, MSI)\end{tabular} &
  PNG &
  \begin{tabular}[c]{@{}c@{}}23\end{tabular} &
  \begin{tabular}[c]{@{}c@{}}1,688\\ (256$\times$256)\end{tabular} &
  \begin{tabular}[c]{@{}c@{}}Pixel-level labels from\\ EMS polygons\end{tabular} \\ \hline
\begin{tabular}[c]{@{}c@{}}GF-FloodNet \\ \citep{zhang2023new} \end{tabular} &
  SAR, MSI &
  Post &
  GeoTiff &
  \begin{tabular}[c]{@{}c@{}}8\end{tabular} &
  \begin{tabular}[c]{@{}c@{}}13,388\\ (256$\times$256)\end{tabular} &
  \begin{tabular}[c]{@{}c@{}}Semi-automatic\\ interactive annotated\\ pixel-level labels\end{tabular} \\ \bottomrule 
\end{tabular}%
}
\end{table}

Second, according to the fusion position or stage, multimodal deep learning architectures can be categorized into early, middle, and late fusion \citep{ma2022crossmodal, park2017rdfnet, qingyun2022cross}. Early fusion integrates data at the input level, middle fusion combines features at intermediate layers, and late fusion merges outputs from separate branches or models at the final stage. Typically, fusion at the input data level is achieved through the channel expansion by concatenating additional data along the channel axis, whereas fusion at the intermediate feature level is accomplished using various feature fusion methods \citep{wang2021geoai}. 

Last, with regard to feature fusion methods, operations such as concatenation, element-wise summation, attention mechanisms, and gating mechanisms are mainly utilized \citep{huang2023deep, mena2024common}. Concatenation and element-wise summation are straightforward operations for fusing features. Concatenation increases the channel dimension by appending features along the channel axis, whereas element-wise summation retains the original dimensionality by summing two features element-wise. However, both concatenation and element-wise summation apply equal weights to features from multiple modalities, ignoring the varying contributions of the features from each modality to the target task \citep{li2020collaborative}. On the other hand, attention and gating mechanisms enable adaptive fusion by adjusting feature importance through learnable weights that emphasize relevant features and suppress less important ones. Specifically, attention mechanisms are generally used to emphasize more relevant features, while gating mechanisms control the flow of information by selectively passing features to optimize the contribution of each one. Furthermore, these operations can be integrated into feature fusion modules specifically designed to address the challenges unique to multimodal deep learning. Previous studies \citep{li2020collaborative, xu2023road} demonstrated that combining multiple feature fusion operations within a modular structure enables the effective utilization of their complementary capabilities.

In post-flood water mapping, prior research on multimodal deep learning has primarily adopted either input-level fusion \citep{bai2021enhancement, konapala2021exploring, wang2024multi} or intermediate feature-level fusion. And the latter has mostly employed concatenation operations \citep{drakonakis2022ombrianet, munoz2021local}. Despite significant advances in multimodal deep learning with geospatial data, adaptive feature fusion methods tailored for post-flood water mapping have undergone limited exploration.

\subsection{Gating Mechanisms in Multimodal Deep Learning with Geospatial Data}
\label{Section2.2}

Gating mechanisms have traditionally been applied in neural networks, such as Long Short-Term Memory (LSTM) networks \citep{hochreiter1997long} and Gated Recurrent Unit (GRU) networks \citep{cho2014learning}, to control the propagation of features. This approach has recently expanded into feature fusion in multimodal deep learning with geospatial data. The implementation of gating mechanisms for feature fusion can be classified based on their approach to gate tensor dimensionality and weighted summation. 

Gate tensor computation follows principles similar to those of attention mechanisms in computer vision, particularly the Convolutional Block Attention Module (CBAM) \citep{woo2018cbam}. Specifically, gate tensors are typically computed as one of the following types: (1) channel-wise \citep{ji2021calibrated, kang2022cfnet, li2020collaborative, zhang2021abmdrnet}, (2) spatial-wise \citep{he2023cross, hosseinpour2022cmgfnet, li2024automatic, zhou2023dsm}, or (3) channel and spatial-wise \citep{cheng2017locality}. For the channel-wise gate vector, either average pooling or max pooling is employed, resulting in a gate vector with dimensions of $\mathbb{R}^{c \times 1 \times 1}$. For the spatial-wise gate map, a convolutional layer with an output channel size of 1 is used, producing the gate map with the shape of $\mathbb{R}^{1 \times h \times w}$.  To generate the channel and spatial-wise gate tensor, a convolutional layer with an output channel size equal to the number of channels in the given feature maps is employed, yielding a gate tensor structured as $\mathbb{R}^{c \times h \times w}$. 

\begin{figure}[t]
  \centering
  \includegraphics[width=1.0 \linewidth]{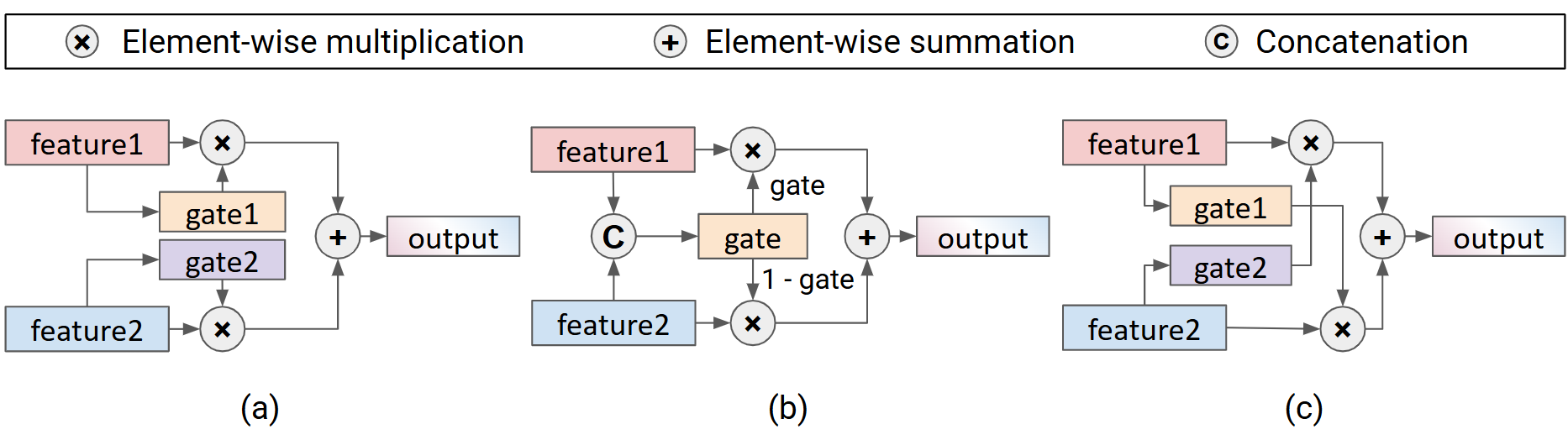}
  \caption{Weighted summation methods in gating mechanisms \citep{huang2023deep, kang2022cfnet}. (a) Independent gating, (b) Complementary gating, and (c) Cross gating.}
  \label{fig1}
\end{figure}

With regard to weighted summation in gating mechanisms, there are primarily three approaches: (1) independent gating, (2) complementary gating, and (3) cross gating (see Fig. \ref{fig1}). In the independent gating approach, two separate gates are leveraged to independently control the contribution of each feature to the fused output. The complementary gating approach, on the other hand, utilizes a single gate and its complement (1 - gate) to ensure that the contributions of the two features are mutually exclusive. Lastly, the cross gating approach employs two gates in a crossed configuration, where each gate controls the contribution of the opposite feature.

Recently, \cite{hosseinpour2022cmgfnet} introduced a gating mechanism that incorporates a spatial-wise gate map with a complementary gating approach for building mapping using RGB bands from satellite data and DSM data. Based on this work, subsequent studies applied the same gating mechanism to different geospatial data modalities in multimodal deep learning. Examples of these studies include flood extent change detection \citep{he2023cross}, urban scene segmentation \citep{zhou2023dsm}, and impervious surface mapping \citep{li2024automatic}.

\subsection{Handling Missing Data in Deep Learning with Multimodal Satellite Data}
\label{Section2.3}

In designing a feature fusion process for multimodal satellite data in deep learning, it is required to consider two key aspects: effectively integrating features across different satellite modalities and robustly handling features extracted from missing data \citep{liu2024multimodal}. Multimodal deep learning models often achieve improved accuracy over unimodal approaches by learning richer feature representations from diverse satellite modalities through feature fusion. However, in practical scenarios, acquiring fully available satellite data for all modalities as input for the model is not always feasible at inference time, due to limitations in data availability \citep{kampffmeyer2018urban, li2021dynamic, liu2024multimodal, wei2023msh}. In such cases, when missing data are not effectively addressed, the performance of a multimodal deep learning model can significantly degrade, potentially yielding worse results than those obtained using a single modality alone \citep{garnot2022multi}. 

In the context of addressing missing data during inference, existing deep learning studies on multimodal satellite data have predominantly explored two scenarios: either assuming complete availability of all modalities \citep{hosseinpour2022cmgfnet, liu2024review, mena2024common} or taking into account availability constraints at the modality-level \citep{adriano2021learning, hong2020more, kampffmeyer2018urban, li2021dynamic, liu2024multimodal, wei2023msh}. Particularly, to address missing data at the modality-level, previous studies have developed novel feature fusion methods \citep{hong2020more} or employed knowledge distillation techniques that leverage learned cross-modal shared representations during inference \citep{kampffmeyer2018urban, li2021dynamic, liu2024multimodal, wei2023msh}. 

In detail, \cite{hong2020more} introduced Cross-Modality Learning (CML), which aims to train a model capable of achieving comparable performance using either a single modality or multiple modalities as input during the inference stage. Their study demonstrated that the cross fusion module effectively balances learned weights across heterogeneous modalities. Additionally, several studies have shown improved performance in handling missing modalities during inference through the hallucination networks based on knowledge distillation \citep{kampffmeyer2018urban, li2021dynamic, wei2023msh}. Building upon these knowledge distillation approaches, \cite{liu2024multimodal} developed a multimodal online knowledge distillation framework that enables inference with either full modalities or any missing modality through simultaneous training of both a modality-fusion network and modality-specific networks. Despite these prior studies addressing modality-level missing data, there has been limited investigation of feature fusion methods designed to handle missing data at the pixel-level. 

\subsection{Weight-shared Decoder in Multimodal Deep Learning}
\label{Section2.4}

Weight sharing in deep learning architectures enables the learning of shared feature representations across different modalities and helps mitigate overfitting by reducing the number of model parameters \citep{ott2020learning}. In multimodal deep learning research, weight-shared architectures have been studied in both encoder and decoder for integrating diverse modalities and leveraging shared feature representations, including text, images, video, and audio data \citep{hickson2022sharing, ngiam2011multimodal, xu2023spnet}. Recently, regarding weight-shared decoders, \cite{hu2021unit} introduced a Unified Transformer (UniT) model, which combines modality-specific encoders with a weight-shared decoder. The UniT model was shown to effectively perform multiple tasks across different domains, including object detection, natural language understanding, and multimodal reasoning, with a compact set of shared parameters in the decoder.

In contrast, research on multimodal deep learning within remote sensing has predominantly focused on weight-shared encoders, specifically through the adoption of Siamese Networks \citep{chopra2005learning}, to extract shared feature representations from different remote sensing data modalities \citep{ge2022improved, lei2022boundary, liu2019siamese, yin2023attention}. Despite the advantages of weight-sharing, investigations into weight-shared decoders in remote sensing remain sparse, with only a few studies exploring this topic \citep{qu2018unsupervised}. This gap emphasizes the necessity for further research on the potential implications of weight-shared decoders in enhancing multimodal deep learning frameworks for satellite data.

\section{Methods}
\label{Section3}

\subsection{Spatially Masked Adaptive Gated Network (SMAGNet)}
\label{Section3.1}

\begin{figure}[t]
  \centering
  \includegraphics[width=1.0 \linewidth]{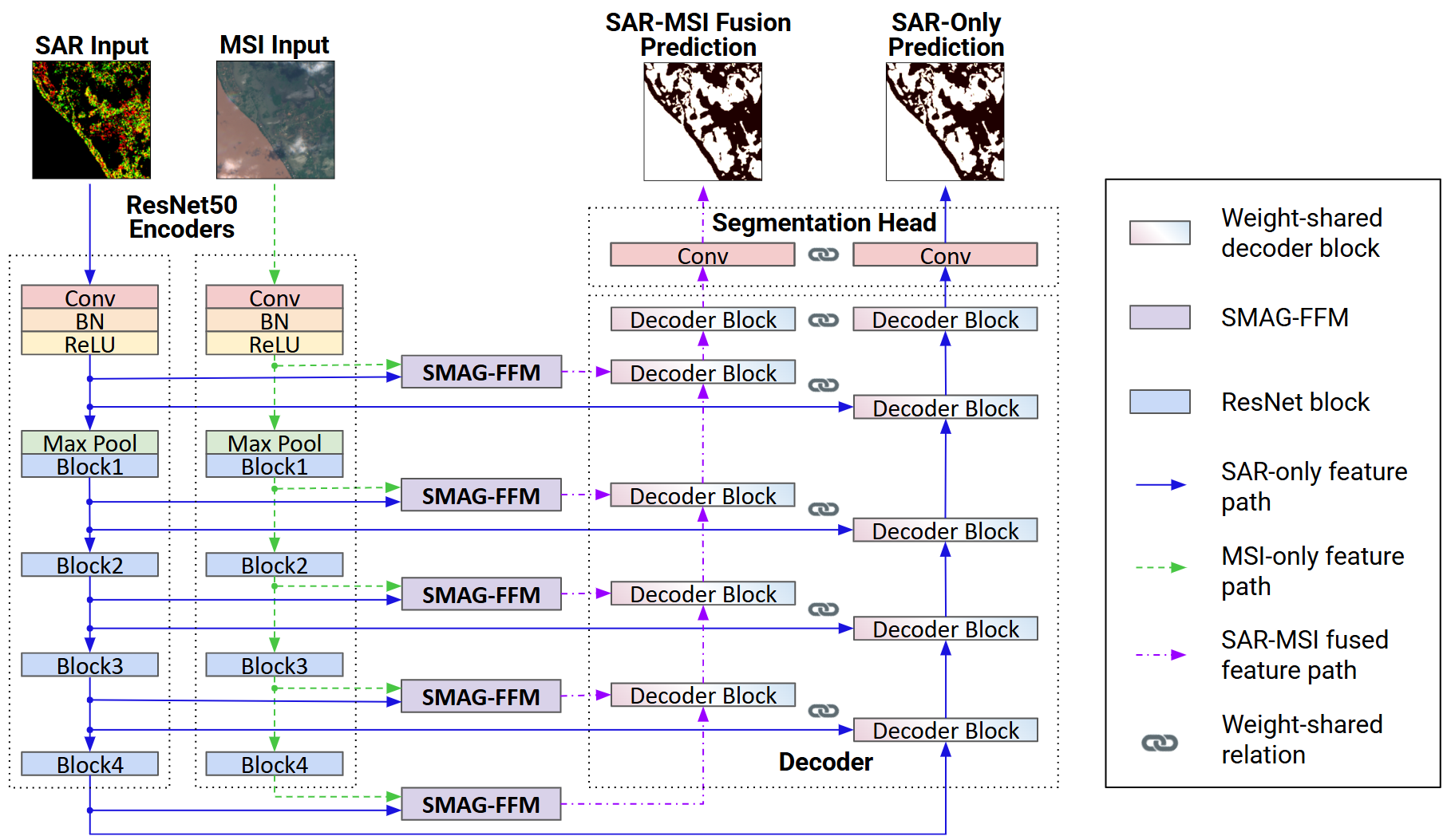}
  \caption{The architecture of Spatially Masked Adaptive Gated Network (SMAGNet).}
  \label{fig2}
\end{figure}

We propose the Spatially Masked Adaptive Gated Network (SMAGNet), a novel multimodal deep learning model aimed at enhancing the accuracy of SAR-based post-flood water mapping during the flood response phase by integrating MSI data. Specifically, this model is designed to utilize SAR data as the primary input and incorporates an adaptive feature fusion mechanism that effectively leverages the complementary features of MSI data and simultaneously addresses the challenges posed by partially or completely missing MSI data. The overall architecture of SMAGNet, as illustrated in Fig. \ref{fig2}, consists of three main components: dual-stream encoders for SAR and MSI data, the Spatially Masked Adaptive Gated Feature Fusion Module (SMAG-FFM), and the weight-shared decoder for post-flood water extent map predictions. 

\subsection{Dual-stream Encoders for SAR and MSI Data}
\label{Section3.2}

SMAGNet employs two separate convolutional neural networks, ResNet50 \citep{he2016deep}, as encoders to extract multi-level features from SAR and MSI data. The ResNet architecture addresses the gradient vanishing problem by introducing skip connections, which directly connect the activations of one layer to subsequent layers, bypassing intermediate layers. These skip connections can be expressed as $F(x) = H(x) - x$, where $H(x)$ represents the function that the network aims to learn, and $x$ denotes the input. By incorporating the identity mapping, the network is tasked with learning the residual $F(x)$ instead of the original mapping $H(x)$, which mitigates the gradient vanishing issue.

In SMAGNet, both ResNet50 encoders progressively produce five feature maps with decreasing spatial dimensions and increasing channel depths. The feature maps from the SAR data are denoted as {$\mathbf{F}^{\text{SAR}}_{(1)}$, $\mathbf{F}^{\text{SAR}}_{(2)}$, $\mathbf{F}^{\text{SAR}}_{(3)}$, $\mathbf{F}^{\text{SAR}}_{(4)}$, $\mathbf{F}^{\text{SAR}}_{(5)}$}, and those from the MSI data are represented as {$\mathbf{F}^{\text{MSI}}_{(1)}$, $\mathbf{F}^{\text{MSI}}_{(2)}$, $\mathbf{F}^{\text{MSI}}_{(3)}$, $\mathbf{F}^{\text{MSI}}_{(4)}$, $\mathbf{F}^{\text{MSI}}_{(5)}$}. At each stage, the spatial dimensions of the feature maps are reduced by a factor of 2, corresponding to \{1/2, 1/4, 1/8, 1/16, and 1/32\} of the original size. The channel dimension of $\mathbf{F}^{\text{SAR}}_{(i)}$ and $\mathbf{F}^{\text{MSI}}_{(i)}$ (where $i$ = 1, 2, 3, 4, 5) is increased across the stages as \{64, 256, 512, 1024, and 2048\}.

\subsection{Spatially Masked Adaptive Gated Feature Fusion Module (SMAG-FFM)}
\label{Section3.3}

\begin{figure}[t]
  \centering
  \includegraphics[width=1.0 \linewidth]{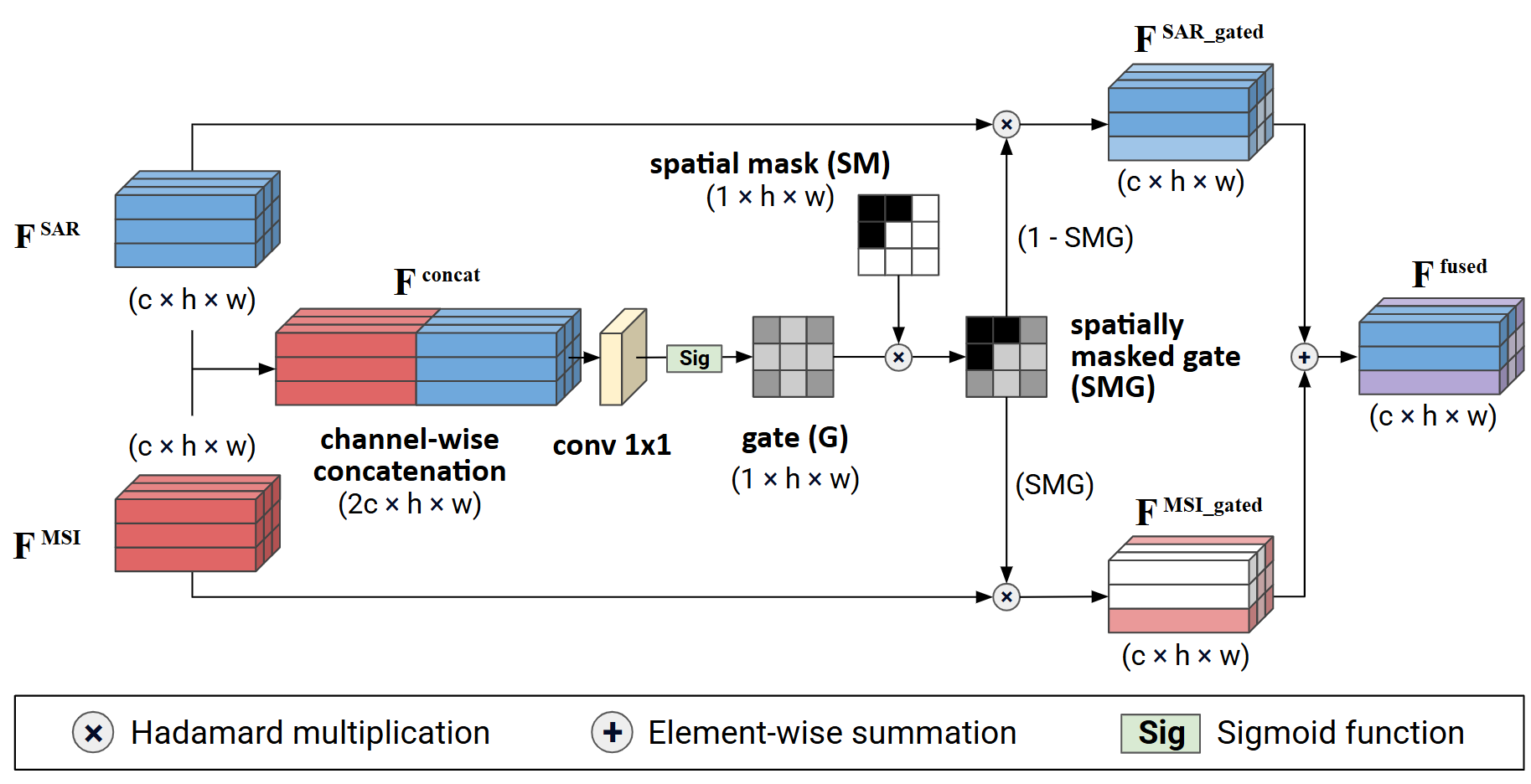}
  \caption{Structure of spatially masked adaptive gated feature fusion module.}
  \label{fig3}
\end{figure}

The multi-level features extracted from dual-stream encoders are fused through the Spatially Masked Adaptive Gated Feature Fusion Module (SMAG-FFM) based on the Spatially Masked Gate (SMG) map. The SMG map is computed utilizing a spatial mask and a spatial-wise gate map. The Fig. \ref{fig3} illustrates the structure of SMAG-FFM to produce the fused feature maps utilizing given two feature maps,  $\mathbf{F}^{\text{SAR}}$ and $\mathbf{F}^{\text{MSI}}$ where both feature maps have the identical spatial dimensional shape of height and weight. In representing feature maps dimensions, $c$ denotes the channel dimension, $h$ represents the height dimension, and $w$ indicates the width dimension. These three components ($c$, $h$, $w$) together define the spatial and channel characteristics of feature maps.

First, $\mathbf{F}^{\text{SAR}}$ and $\mathbf{F}^{\text{MSI}}$ are concatenated along the channel dimension, forming a combined feature maps $\mathbf{F}^{\text{concat}}$ with twice the original number of channels (see Eq. \ref{eq1}). The concatenated feature maps are then passed through a 1x1 convolutional layer, followed by a sigmoid activation function, to produce a spatial-wise gate map $\mathbf{G}$ (see Eq. \ref{eq2}). This approach, which combines a spatial-wise gate map with a complementary gating mechanism, has demonstrated effectiveness in prior feature fusion studies \citep{he2023cross, hosseinpour2022cmgfnet, li2024automatic, woo2018cbam, zhou2023dsm}. 

\begin{equation}
\label{eq1}
    \mathbf{F}^{\text{concat}} = \left[\mathbf{F}^{\text{SAR}}; \mathbf{F}^{\text{MSI}}\right]
    \text{ ;} \quad \mathbf{F}^{\text{concat}} \in \mathbb{R}^{2c \times h \times w}.
\end{equation}

\begin{equation}
\label{eq2}
    \mathbf{G} = \sigma (\text{conv}_{1\times1} (\mathbf{F}^{\text{concat}})) \text{ ;} \quad \mathbf{G} \in \mathbb{R}^{1 \times h \times w}.
\end{equation}

Subsequently, a Spatial Mask (SM) designed to handle missing data is applied to the gate map $\mathbf{G}$ through element-wise multiplication, yielding a SMG map (see Eq. \ref{eq3}). Masking has traditionally been an effective method for filtering out unnecessary information, and previous studies have applied masking to address missing values in deep learning, particularly in time series data analysis \citep{che2018recurrent}. In SMAGNet, a spatial mask is generated from the missing data pixels in the MSI data by downsampling them to match the dimensions of the gate map $\mathbf{G}$.

\begin{equation}
\label{eq3}
    \mathbf{SMG} = \mathbf{SM} \otimes \mathbf{G} \text{ ;} \quad \mathbf{SMG} \in \mathbb{R}^{1 \times h \times w}.
\end{equation}


Afterward, through Hadamard (element-wise) multiplication, the SMG map is applied to $\mathbf{F}^{\text{MSI}}$, whereas the complemented SMG map, (1 - SMG), is applied to $\mathbf{F}^{\text{SAR}}$ (see Eq. \ref{eq4} and \ref{eq5}). The SMG map modulates the emphasis on each data source by adaptively adjusting their contribution. Specifically, in regions where the SMG map values exceed 0.5, the predictions are more influenced by the MSI features. On the other hand, in areas where values in the SMG map fall below 0.5, the SAR features become the primary contributors to the predictions. This adaptive weighting scheme harnesses the complementary strengths of SAR and MSI data by selectively fusing features based on the spatial context from the SMG map, thereby improving prediction accuracy through the fusion process. Finally, the two weighted feature maps, $\mathbf{F}^{\text{SAR\_gated}}$ and $\mathbf{F}^{\text{MSI\_gated}}$, are combined through element-wise summation to produce the final fused feature maps $\mathbf{F}^{\text{fused}}$ (see Eq. \ref{eq6}).

\begin{equation}
\label{eq4}
    \mathbf{F}^{\text{MSI\_gated}} = \mathbf{F}^{\text{MSI}} \otimes \mathbf{SMG} 
    \text{ ; } \mathbf{F}^{\text{MSI\_gated}} \in \mathbb{R}^{c \times h \times w}.
\end{equation}

\begin{equation}
\label{eq5}
    \mathbf{F}^{\text{SAR\_gated}} = \mathbf{F}^{\text{SAR}} \otimes (1 - \mathbf{SMG}) 
    \text{ ; } \mathbf{F}^{\text{SAR\_gated}} \in \mathbb{R}^{c \times h \times w}.
\end{equation}

\begin{equation}
\label{eq6}
    \mathbf{F}^{\text{fused}} = \mathbf{F}^{\text{SAR\_gated}} \oplus \mathbf{F}^{\text{MSI\_gated}} 
    \text{ ; } \mathbf{F}^{\text{MSI\_gated}} \in \mathbb{R}^{c \times h \times w}.
\end{equation}

In SMAG-FFM, when missing data pixels are present in the MSI data, the spatial mask functions to assign a lower gating weight to the channel-wise feature vector, $\mathbf{f}^{\text{MSI}}_{i \text{,} j}$, in proportion to the amount of missing pixels within each area corresponding to the position of the spatial mask ($i$, $j$). Furthermore, if missing data pixels cover an entire area corresponding to a single pixel located at ($i$, $j$) in the spatial mask, the SMG value for that pixel becomes zero. Consequently, the spatial mask operates to preserve the feature vector at ($i$, $j$) in $\mathbf{F}^{\text{SAR}}$, $\mathbf{f}^{\text{SAR}}_{i \text{,} j}$, during the feature fusion process when missing data pixels are present in the MSI data (see Eq. \ref{eq7}).


\begin{equation}
\label{eq7}
\begin{aligned}
    &\mathbf{f}^{\text{fused}}_{i,j} = \begin{cases}
    \mathbf{f}_{i,j}^{\text{SAR}}, \quad \text{if } \text{SMG}_{i,j} = 0 \\[2ex]
    \mathbf{f}_{i,j}^{\text{SAR}} \otimes (1 - \text{SMG}_{i,j}) \oplus  \mathbf{f}_{i,j}^{\text{MSI}} \otimes \text{SMG}_{i,j}, \text{otherwise,}
    \end{cases}
\end{aligned}
\end{equation}
where $\mathbf{f}_{i,j}^{\text{fused}}$, $\mathbf{f}_{i,j}^{\text{SAR}}$, $\mathbf{f}_{i,j}^{\text{MSI}} \in \mathbb{R}^c$.

\subsection{Weight-shared Decoder}
\label{Section3.4}

The SMAG-FFM outputs fused feature maps that spatially contain either SAR-MSI fused feature vectors or SAR-only feature vectors, depending on the presence of missing data pixels in the given MSI data. To train both SAR-MSI fused features and SAR-only features within a unified decoder, SMAGNet employs a weight-shared decoder. Specifically, by sharing weights in convolutional layers across features extracted from different modalities, SMAGNet enables straightforward pixel-level shared feature representation learning, in contrast to knowledge distillation methods that are commonly employed for modality-level shared representation learning \citep{kampffmeyer2018urban, li2021dynamic, wei2023msh}.  

\begin{figure}[t]
  \centering
  \includegraphics[width=0.5 \linewidth]{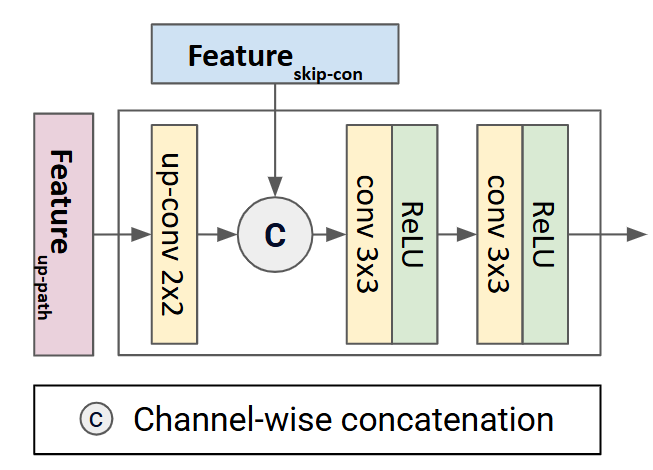}
  \caption{Decoder block structure in SMAGNet.}
  \label{fig4}
\end{figure}

In SMAGNet, the weight-shared decoder processes dual feature paths using identical weights at each layer: one path for SAR-MSI fused features and the other for SAR-only features (see Fig. \ref{fig2}). The weight-shared decoder consists of 5 decoder blocks with dimensions of \{256, 128, 64, 32, 16\} respectively. In addition, as illustrated in Fig. \ref{fig4}, the decoder block at each stage includes two consecutive convolutional layers, each paired with a Rectified Linear Unit (ReLU) activation function. This decoder block structure is the same as the decoder block in the U-Net \citep{ronneberger2015u} model. The output feature maps from the previous stage passes through an upsampling convolutional (up-conv) layer, which increases its spatial dimensions to match those of the skip connection feature. Then, the output feature maps from the up-conv layer and the skip connection feature maps are concatenated along the channel dimension.


As the outputs of the weight-shared decoder, two distinct feature paths generate separate prediction outputs. Each output is evaluated based on the Binary Cross-Entropy (BCE) loss function with labeled data derived from SAR data (see Eq. \ref{eq8}). The final loss function is obtained by summing equally weighted BCE loss terms ($w$ = 0.5) to train the two feature paths in a balanced manner (see Eq. \ref{eq9}).


\begin{equation}
\label{eq8}
    \begin{aligned}
        \mathbf{L}_{\text{BCE}} &(\hat{\mathbf{Y}}, \mathbf{Y}) = \frac{1}{n} \sum_{i=0}^n \Big(y_i \times \log (\hat{y}_i) + (1 - y_i) \times \log (1 - \hat{y}_i)\Big).
    \end{aligned}
\end{equation}

\begin{equation}
\label{eq9}
    \begin{aligned}
        \mathbf{L} = w \times \mathbf{L}_{\text{BCE}}(\hat{\mathbf{Y}}_{\text{SAR}}, \mathbf{Y}) + (1-w) \times \mathbf{L}_{\text{BCE}}(\hat{\mathbf{Y}}_{\text{fused}}, \mathbf{Y}).
    \end{aligned}
\end{equation}

\section{Experimental Setup}
\label{Section4}

\subsection{Dataset}
\label{Section4.1}

The proposed method was evaluated using the C2S-MS (Cloud to Street-Microsoft) Floods dataset \citep{Cloud_to_Street2022}, an AI-ready dataset suited for SAR-based post-flood water extent mapping with complementary feature fusion of MSI data. In reviewing additional publicly available benchmark datasets for our experiments, we found no others that were adequately suited to our study objectives. To the best of our knowledge, the C2S-MS Floods dataset is unique in providing manually annotated labels based directly on SAR data for multimodal deep learning in post-flood water mapping. In contrast, the other datasets primarily relied on MSI data for labeling their data. For instance, cloud-covered areas are annotated as missing data in the labeled data of the SenFloods11 dataset \citep{bonafilia2020sen1floods11}, and the GF-FloodNet dataset \citep{zhang2023new} employs semi-automated MSI-based annotation.

The C2S-MS Floods dataset contains 900 paired SAR and MSI images, each with a size of 512 $\times$ 512 pixels, from 18 global flood events. The SAR data, acquired from Sentinel-1, includes two polarization bands, VV (Vertical transmit, Vertical receive) and VH (Vertical transmit, Horizontal receive), and the MSI data, obtained from Sentinel-2, provides 13 spectral bands. Both types of satellite data were acquired over the same locations within four days after the flood events that occurred between 2016 and 2020. SAR data was pre-processed with orbit correction, noise removal, calibration, terrain correction, and conversion to decibels  \citep{Cloud_to_Street2022}. In addition, SAR and MSI data were both resampled to 10m resolution for all bands. In terms of MSI data availability, approximately 11\% of the MSI data in the C2S-MS Floods dataset contains missing data pixels with varying proportions.

\begin{figure}[t]
  \centering
  \includegraphics[width=1.0 \linewidth]{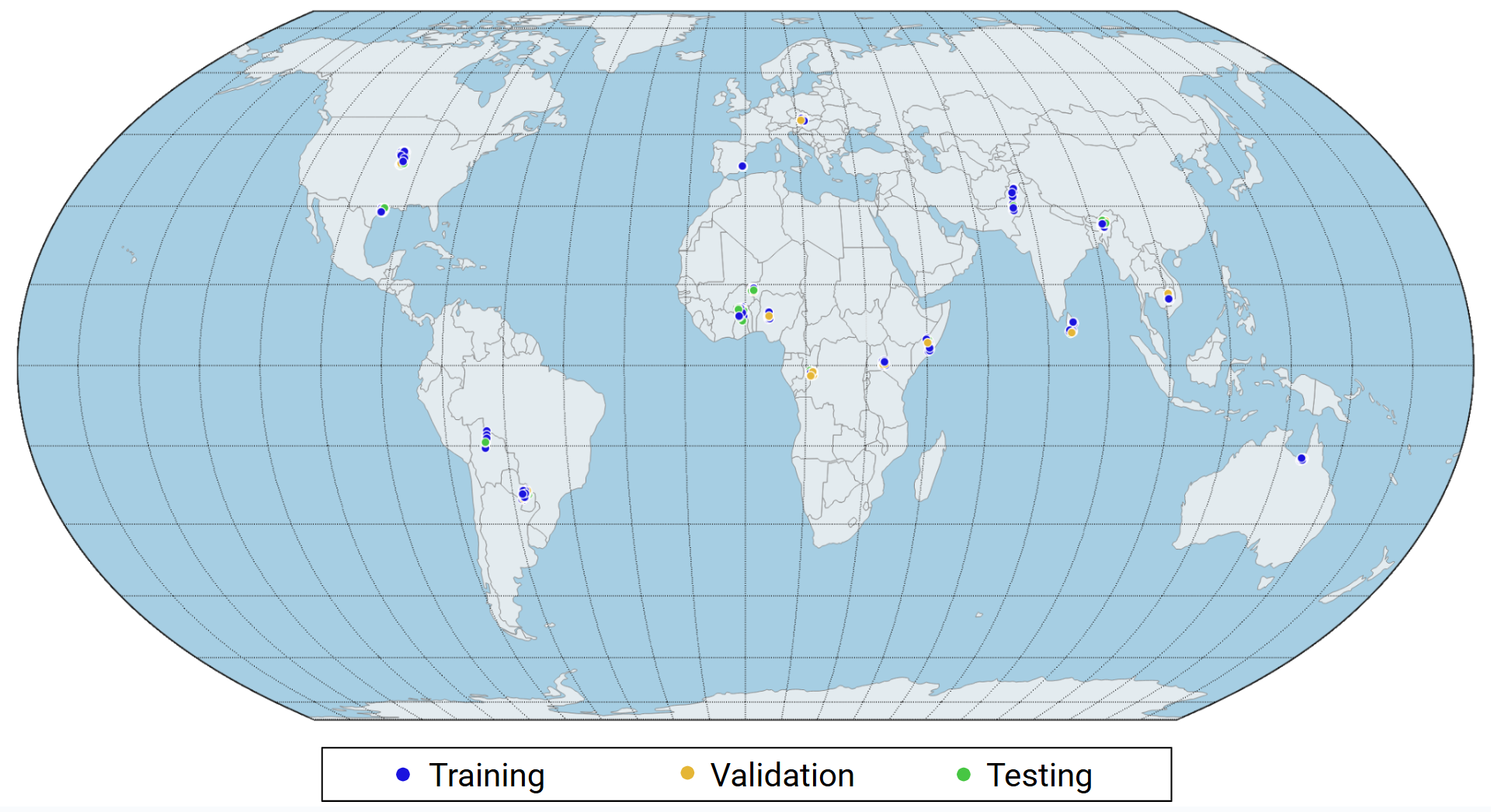}
  \caption{Spatial distribution of C2S-MS Floods dataset.}
  \label{fig5}
\end{figure}

For input bands, two bands (VV and VH) were selected from SAR data and four bands (Red, Green, Blue, and NIR), which have an original spatial resolution of 10m, were chosen from MSI data. Other MSI bands, which originally had spatial resolutions of 20m or greater, were excluded from the input. In addition, stratified random sampling based on acquisition location was performed to split the data into training, validation, and test datasets in a 6:2:2 ratio. For efficient GPU memory utilization, we set the input data resolution to 256 $\times$ 256. Therefore, each image in the validation and test datasets was divided into four non-overlapping 256 $\times$ 256 patches, resulting in 720 data samples in both validation and test dataset. To achieve consistent scaling of the data across all samples, band-wise normalization was applied to each spectral band individually using the mean and standard deviation calculated from the training dataset. The spatial distribution of the C2S-MS Floods dataset across training, validation, and test splits is illustrated in Fig. \ref{fig5}.


\subsection{Implementation Details}
\label{Section4.2}

All models were implemented using the PyTorch framework, and all experiments were conducted on a workstation with an NVIDIA RTX A5000 and 251 GB of memory under the same experimental parameter conditions. In the model training, the Adam \citep{kingma2014adam} optimizer was used, with the weight decay set to 0.0 and the initial learning rate set to 5e-4. The batch size and number of epochs were 16 and 200, respectively. For data augmentation, random crop and random flip were applied in all experiments. As a loss function, binary cross entropy was employed. Specifically in SMAGNet, for the SAR data encoder, the weights were randomly initialized, while the weights for the MSI data encoder were initialized using pre-trained weights from ImageNet \citep{deng2009imagenet}.

The final model was selected as the one that achieved the lowest validation loss during the training. Then, the optimal threshold for classifying each pixel as either flood or non-flood was determined by identifying the value on the Precision-Recall curve that maximized the Intersection over Union (IoU) score, based on the validation dataset. This determined threshold was subsequently applied to the predictions made on the test dataset to evaluate the overall model performance.

\subsection{Evaluation Metrics}
\label{Section4.3}

In this study, four evaluation metrics were used to measure the model performance for post-flood water mapping: Overall Accuracy (OA), Precision, Recall, and Intersection over Union (IoU). These metrics are calculated based on the True Positives (TP), False Positives (FP), False Negatives (FN), and True Negative (TN) from the confusion matrix. When interpreting the prediction outcomes, false positives (FPs) are considered as a kind of over-detection, referring to pixels that are not annotated as flood in the labeled data, but are predicted as flood pixels by the model. Conversely, false negatives (FNs) are considered under-detected pixels, which are annotated as flood but predicted as non-flood pixels. OA is defined as the proportion of correctly predicted pixels out of the total number of pixels, providing a straightforward measure of classification accuracy. However, since real-world datasets such as post-flood water extent maps frequently exhibit class imbalances, OA may not provide a reliable assessment of the model performance. To complement this limitation of OA, Precision, Recall, and IoU are additionally employed. Precision represents the percentage of correctly predicted positive pixels (TP) among all pixels predicted as positive (TP + FP). Recall measures the percentage of correctly predicted positive pixels (TP) out of all actual positive pixels in the ground truth (TP + FN). IoU, a metric that evaluates the overlap between the predicted segmentation and the ground truth, is calculated as the ratio of the intersection to the union of the two sets.

\begin{equation}
\label{eq10}
    \text{OA} = \frac{\text{TP} + \text{TN}}{\text{TP} + \text{TN} + \text{FN} + \text{FP}} \text{.}
\end{equation}

\begin{equation}
\label{eq11}
    \text{Precision} = \frac{\text{TP}}{\text{TP} + \text{FP} } \text{.}
\end{equation}

\begin{equation}
\label{eq12}
    \text{Recall} = \frac{\text{TP}}{\text{TP} + \text{FN} } \text{.}
\end{equation}

\begin{equation}
\label{eq13}
    \text{IoU} = \frac{\text{TP}}{\text{TP} + \text{FN} + \text{FP} } \text{.}
\end{equation}

\subsection{Comparison Methods}
\label{Section4.4}

To assess the performance of the proposed method, we compared SMAGNet with several classic and state-of-the-art multimodal deep learning approaches for semantic segmentation. However, since not all comparison models were originally designed to process both SAR and MSI inputs, detailed modifications were made for the experiments. Specifically, U-Net \citep{ronneberger2015u}, U-Net++ \citep{zhou2019unetplusplus}, PSPNet \citep{zhao2017pyramid}, DeepLabV3+ \citep{chen2018encoder}, FPN \citep{lin2017feature} were used with a ResNet50 encoder, and each model was configured to process SAR and MSI input data through channel expansion. For multimodal deep learning models that utilize RGB band and DSM or depth data, such as FuseNet \citep{hazirbas2017fusenet}, VFuseNet \citep{audebert2018beyond}, CMFNet \citep{ma2022crossmodal}, CMGFNet \citep{hosseinpour2022cmgfnet}, FTransUNet \citep{ma2024multilevel}, SAR data was employed as the main input data instead of RGB data, and MSI data was used as the supplementary input data instead of DSM or depth data. The following provides detailed descriptions of the deep learning models used in the experiments. 

\begin{enumerate}
    \renewcommand{\labelenumi}{\arabic{enumi})}  
    \item U-Net \citep{ronneberger2015u} is a convolutional neural network composed of an encoder, decoder, and skip connections, widely used for image segmentation tasks. The decoder block structure is identical to the one used in SMAGNet.
    \item U-Net++ \citep{zhou2019unetplusplus} is an extension of the U-Net model that introduces nested and dense skip connections to improve segmentation performance. 
    \item PSPNet \citep{zhao2017pyramid} is a deep learning-based semantic segmentation model that efficiently captures global context by combining multi-scale contextual information through pyramid pooling. 
    \item DeepLabV3+ \citep{chen2018encoder} is an improved version of the DeepLab model that combines atrous spatial pyramid pooling (ASPP) with an encoder-decoder structure. 
    \item FPN \citep{lin2017feature} is the Feature Pyramid Network that uses a bottom-up pathway to extract multi-scale feature maps and a top-down pathway with lateral connections to refine and merge these features at different resolutions. 
    \item FuseNet \citep{hazirbas2017fusenet} is a multimodal fusion network for semantic segmentation that simultaneously extracts features from RGB and depth images, fusing depth information into the RGB feature maps progressively as the network deepens. 
    \item VFuseNet \citep{audebert2018beyond} is an extension of FuseNet that modifies the original asymmetrical architecture to a symmetrical architecture, eliminating the need to determine a main input data source.
    \item FTransUNet \citep{ma2024multilevel} is a multimodal fusion model for semantic segmentation that integrates a convolutional neural network and a transformer to effectively fuse shallow and deep-level features for accurate local detail and global semantic representation. 
    \item CMGFNet \citep{hosseinpour2022cmgfnet} is a cross-modal gated fusion network designed to extract building footprints from very high-resolution remote sensing images and digital surface models by employing separate encoders for RGB and DSM data, integrating features through a gated fusion module and a multi-level feature fusion strategy. 
    \item CMFNet \citep{ma2022crossmodal} is a crossmodal multiscale fusion network that leverages transformer architecture to fuse multiscale features from optical remote sensing images and DSM data using cross-attention mechanisms.
    \item MCANet \citep{li2022mcanet} is a multimodal-cross attention network designed for land use classification by fusing optical and SAR images, utilizing independent feature extraction, second-order hidden feature mining, and multi-scale feature fusion.
    \item MFGFUnet \citep{wang2024multi} is a multi-modality fusion network with a gated multi-filter inception module and Gated Channel Transform (GCT) \citep{yang2020gated} skip connections, designed to enhance water area segmentation.
    
\end{enumerate}

\section{Results}
\label{Section5}

\subsection{Comparative Study}
\label{Section5.1}

The comparative study aims to evaluate the performance of SMAGNet compared with other deep learning models based on a multimodal approach, using four metrics described in Section \ref{Section4.3}. For reliable performance evaluation, each experiment was conducted 10 times, and we reported the mean and standard deviation of each metric. Table \ref{tab2} presents the experimental results of the comparative study. 

As a baseline for SAR-based post-flood water mapping, we used a U-Net model trained solely on SAR data, referred to as U-Net (SAR). All deep learning models based on a multimodal approach exhibited superior performance to the U-Net (SAR) across all four metrics. This observation aligns with the findings of previous research \citep{konapala2021exploring} and highlights the effectiveness of multimodal deep learning in post-flood water mapping. Notably, SMAGNet outperformed other multimodal deep learning models by achieving the highest scores in three of the four metrics: 86.47\% for IoU, 92.45\% for Recall, and 97.73\% for Accuracy. For Precision, SMAGNet achieved the second-best score at 93.05\%, with CMGFNet achieving the highest at 94.85\%.

\begin{table}[t]
\centering
\caption{Experimental results of the comparative study.}
\label{tab2}
\resizebox{\textwidth}{!}{%
\begin{tabular}{lcccc}
\toprule 
\multicolumn{1}{c}{Model} & IoU (\%)      & Precision (\%) & Recall (\%)   & OA (\%)       \\ \hline\hline
U-Net (SAR)               & 79.65 (±0.96) & 90.81 (±0.83)  & 86.64 (±1.03) & 96.52 (±0.18) \\ \hline
PSPNet                    & 82.65 (±0.85) & 90.83 (±0.93)  & 90.19 (±1.29) & 97.02 (±0.15) \\ \hline
VFuseNet                  & 83.33 (±1.00) & 92.98 (±0.62)  & 88.92 (±0.89) & 97.20 (±0.18) \\ \hline
FuseNet                   & 83.40 (±1.13) & 92.95 (±0.71)  & 89.03 (±0.87) & 97.21 (±0.20) \\ \hline
FTransUNet                & 83.93 (±2.64) & 92.19 (±1.08)  & 90.34 (±2.46) & 97.28 (±0.47) \\ \hline
FPN                       & 84.25 (±0.96) & 91.10 (±1.03)  & 91.80 (±0.54) & 97.30 (±0.19) \\ \hline
U-Net++                   & 84.41 (±1.54) & 92.75 (±0.69)  & 90.36 (±1.32) & 97.37 (±0.27) \\ \hline
DeepLabV3+                & 84.48 (±1.19) & 92.04 (±0.68)  & 91.14 (±1.26) & 97.37 (±0.21) \\ \hline
CMGFNet        & 84.70 (±0.59)          & \textbf{94.85} (±0.46) & 88.78 (±0.71)          & 97.48 (±0.10)          \\ \hline
CMFNet                    & 84.95 (±0.87) & 92.31 (±0.93)  & 91.43 (±0.95) & 97.45 (±0.16) \\ \hline
U-Net                     & 84.96 (±0.97) & 92.88 (±0.60)  & 90.88 (±0.92) & 97.47 (±0.17) \\ \hline
MCANet                    & 85.48 (±0.99) & 92.47 (±0.78)  & 91.87 (±0.82) & 97.54 (±0.18) \\ \hline
MFGFUnet                  & 85.96 (±0.57) & 92.84 (±0.98)  & 92.07 (±0.73) & 97.63 (±0.11) \\ \hline
SMAGNet (Ours) & \textbf{86.47} (±0.61) & 93.05 (±0.76)          & \textbf{92.45} (±0.83) & \textbf{97.73} (±0.11) \\ \bottomrule 
\end{tabular}%
}
\end{table}

Specifically, in terms of IoU, SMAGNet achieved the highest performance, followed by MFGFUnet (85.96\%) and MCANet (85.48\%), both of which are intended to utilize SAR and MSI data as input. Following in IoU scores were U-Net (84.96\%), CMFNet (84.95\%), and CMGFNet (84.70\%). CMFNet and CMGFNet are multimodal deep learning architectures specifically designed to leverage optical satellite imagery and DSM data. In addition, SMAGNet showed comparable performance variability to other multimodal deep learning models across four evaluation metrics, with standard deviations of ±0.61\% for IoU, ±0.76\% for Precision, ±0.83\% for Recall, and ±0.13\% for Accuracy. As a result, these experimental results demonstrate that SMAGNet not only achieved superior performance in most metrics but also maintained stability comparable to that of other models across repeated experiments.

\begin{figure}[p]
  \centering
  \includegraphics[width=1.0 \linewidth]{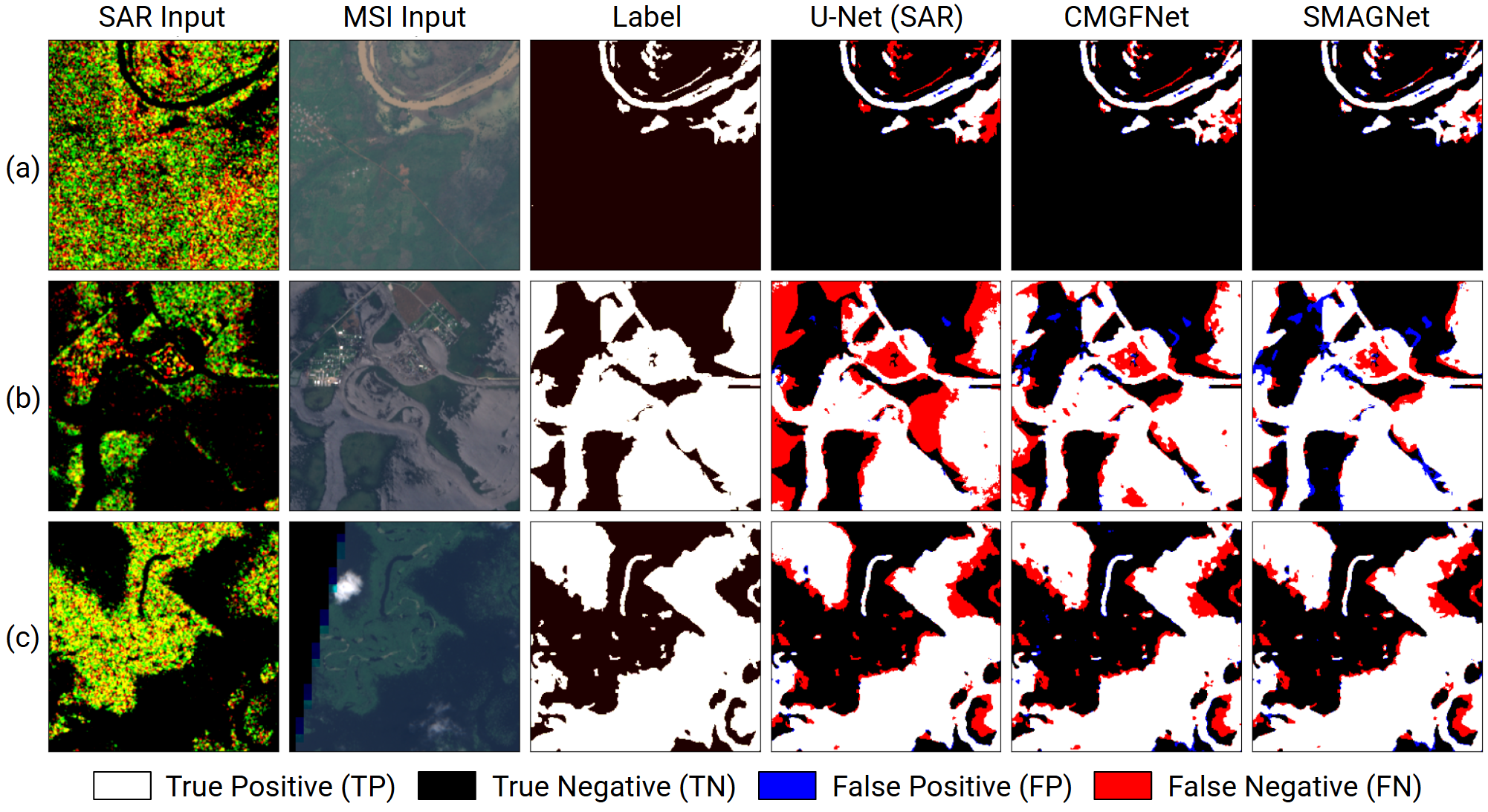}
  \caption{Visualizations of sample prediction results from U-Net (SAR), CMGFNet, and SMAGNet. U-Net (SAR) is the baseline, CMGFNet achieved the highest Precision, and SMAGNet achieved the highest IoU, Recall, and OA.}
  \label{fig6}
\end{figure}

\begin{figure}[p]
  \centering
  \includegraphics[width=0.7 \linewidth]{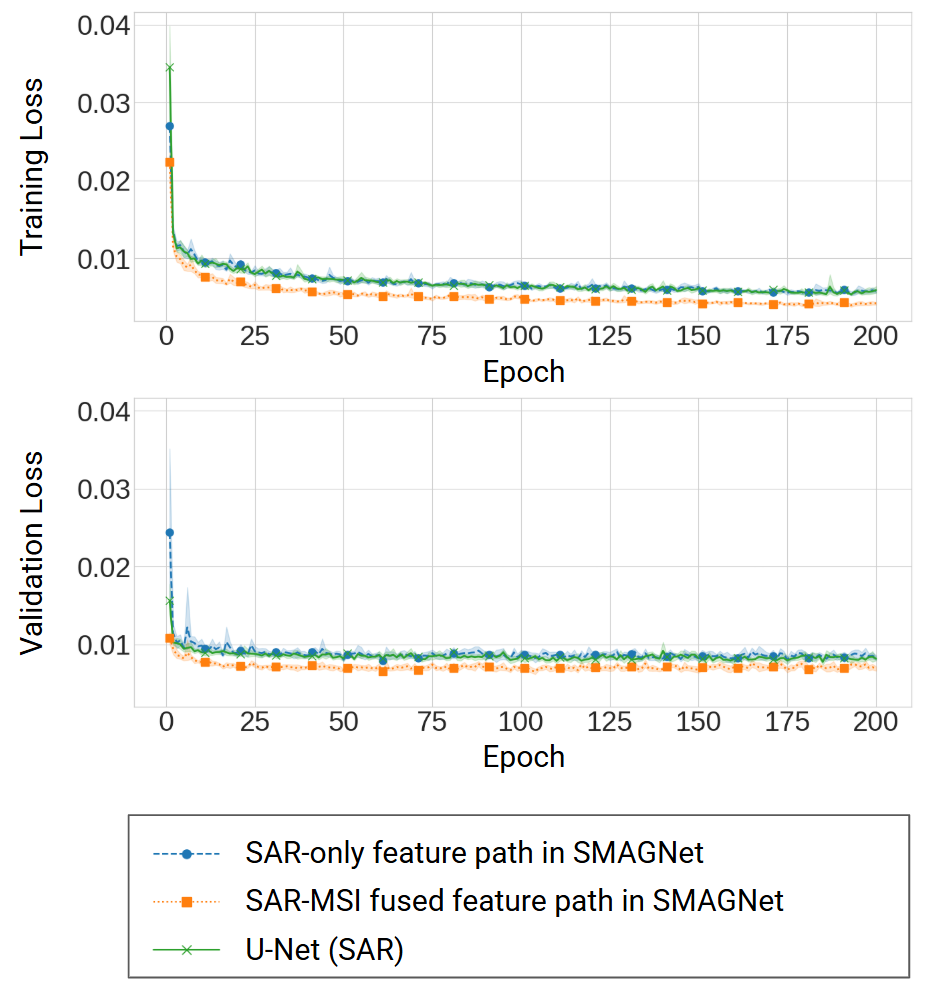}
  \caption{Comparison of the training and validation loss curves between SMAGNet and U-Net (SAR). The line plot shows the loss for each epoch, with markers added every 10 epochs for visual distinction.}
  \label{fig7}
\end{figure}

The visualization results in Fig. \ref{fig6} (a) closely align with the quantitative evaluations in Table \ref{tab2}. Specifically, compared to the U-Net (SAR), both CMGFNet and SMAGNet visually showed fewer misclassified pixels. Particularly, SMAGNet, which achieved the highest Recall, exhibited the fewest false negatives (e.g., under-detection), whereas CMGFNet, with the highest Precision, showed fewer false positives (e.g., over-detection). In the case of Fig. \ref{fig6} (b), although U-Net (SAR) was well trained, as indicated by the converging training and validation loss curves in Fig. \ref{fig7}, some samples exhibited markedly larger misclassified pixels. In contrast, using the same input data, models that incorporate MSI data showed a noticeable reduction in misclassified pixels. Fig. \ref{fig6} (c) shows the visualization result for a case in which part of the MSI data is missing. Compared to U-Net (SAR) and CMGFNet, SMAGNet visually exhibits fewer false negatives in areas where the MSI data is missing.

To more thoroughly investigate the performance improvement achieved through the incorporation of MSI data into SMAGNet, we utilized histograms to analyze the number of misclassified pixels with respect to the Normalized Difference Vegetation Index \citep[NDVI; ][]{townshend1986analysis, tucker1986satellite} and Near-Infrared (NIR) reflectance across the entire test dataset. The histograms were then compared against the results from U-Net (SAR). NDVI is a widely adopted indicator for quantifying vegetation density. Specifically, negative NDVI values typically indicate the presence of clouds or water, values near zero correspond to bare soil, and positive values represent vegetation cover. Therefore, NDVI can be utilized to characterize the misclassified pixels by the two models in areas with flooded vegetation. For this purpose, in this study, NDVI values between 0.1 and 0.5 were interpreted as indicating sparsely vegetated areas, while values above 0.5 were considered to represent densely vegetated areas. In addition, the NIR reflectance is effective for identifying water-covered areas due to its sensitivity to surface water and low reflectance caused by water absorption. However, cloud shadows also exhibit low NIR reflectance \citep[typically below 0.1; ][]{feyisa2014automated}, which can result in false positives by causing non-water areas to be misclassified as water. Therefore, by quantitatively comparing the number of misclassified pixels between the two models (SMAGNet and U-Net (SAR)) in terms of NDVI and NIR reflectance, we assessed the contribution of integrating MSI data into post-flood water mapping in different environmental conditions.

\begin{figure}[t]
  \centering
  \includegraphics[width=1.0 \linewidth]{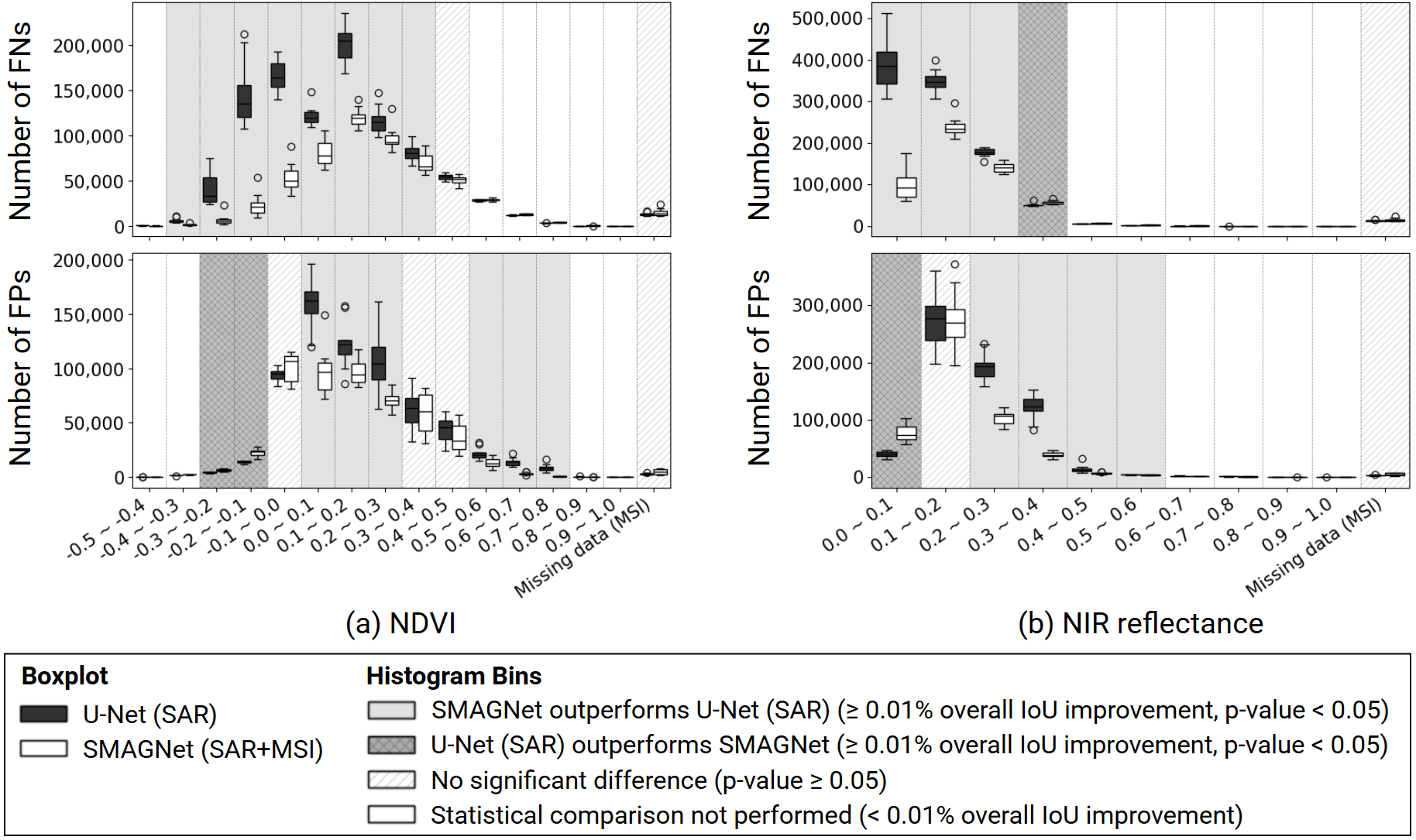}
  \caption{Histograms of the number of misclassified pixels with respect to Normalized Difference Vegetation Index (NDVI) and Near-InfraRed (NIR) reflectance across the entire test dataset.}
  \label{fig8}
\end{figure}

Fig. \ref{fig8} shows the histograms of misclassified pixels by SMAGNet and U-Net (SAR) based on NDVI and NIR reflectance. In terms of IoU score, the gray background denotes the range where SMAGNet exhibits a statistically significant improvement over U-Net (SAR), whereas the cross-hatched background indicates the range where U-Net (SAR) outperforms SMAGNet with statistical significance. The diagonally hatched background represents a range where there is no statistically significant difference between the two models. The white background indicates the range where the difference in the number of misclassified pixels is too small (below 0.01\%) to significantly affect the IoU score; therefore, statistical comparison is not performed.

In Fig. \ref{fig8} (a), SMAGNet significantly reduced false negatives in the NDVI range from -0.4 to 0.4 compared to U-Net (SAR). This indicates that SMAGNet improves post-flood water detection in areas such as water bodies, bare soil, and sparse vegetation. However, in areas with dense vegetation (NDVI above 0.5), the difference in false negatives between SMAGNet and U-Net (SAR) was negligible, corresponding to less than a 0.01\% difference in the IoU score. On the other hand, SMAGNet effectively decreased false positives in the NDVI ranges from 0.0 to 0.3 and from 0.5 to 0.8, indicating enhanced precision in vegetated areas, including dense vegetation. Notably, the false positives of SMAGNet were slightly higher than those of U-Net (SAR) in the NDVI range from -0.3 to -0.1, which may reflect the substantial reduction in false negatives observed in the same range. Overall, the incorporation of MSI data improves post-flood water mapping performance across most NDVI ranges, except in densely vegetated areas where false negatives remain comparable to those of U-Net (SAR).

In Fig. \ref{fig8} (b), SMAGNet substantially reduced false negatives in the NIR reflectance range from 0 to 0.1, indicating improved post-flood water detection at low NIR reflectance values. Although false positives slightly increased in this range, this may be due to enhanced sensitivity (or recall). The increase in false positives at NIR reflectance values between 0 and 0.1 could lead to more misclassifications in regions such as cloud shadows. Nonetheless, considering both the reduction in false negatives and the slight increase in false positives at NIR reflectance values between 0 and 0.1, the incorporation of MSI data led to a clear performance improvement. Notably, for pixels with missing values in MSI data, both SMAGNet and U-Net (SAR) exhibited a statistically comparable level of misclassified pixels, including both false negatives and false positives.

\subsection{Robustness Study}
\label{Section5.2}

\begin{figure}[b]
  \centering
  \includegraphics[width=1.0 \linewidth]{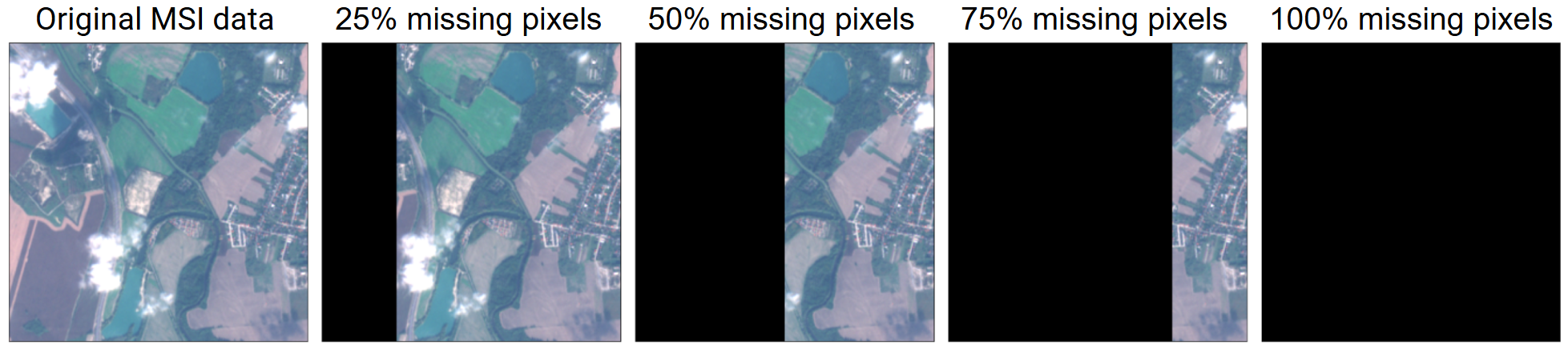}
  \caption{Visualization of sample MSI data that are replaced with missing data pixels at proportions of 25\%, 50\%, 75\%, and 100\%.}
  \label{fig9}
\end{figure}

In the robustness study, we designed an experiment to assess the effectiveness of SMAGNet in addressing missing data pixels in MSI data for enhanced SAR-based post-flood water mapping. To achieve this experimental objective, we replaced the original MSI data in the test dataset with missing data pixels at proportions of 25\%, 50\%, 75\%, and 100\%, as shown in Fig. \ref{fig9}. The missing data pixels are represented as black regions, progressively covering larger portions of the image from left to right as the replacement ratio increases. The robustness experiments were conducted using the same trained models as the comparative study in Section \ref{Section5.1}, with the only modification being the use of MSI data in the test dataset where pixels were replaced by missing data at specific percentages.

Table \ref{tab3} presents the experimental results of the robustness study, showing a noticeable decline in IoU scores across all models as the proportion of missing data increased. However, the extent of performance degradation varied significantly depending on the models. SMAGNet, in particular, exhibited the highest level of robustness, consistently achieving the top IoU scores across all levels of missing data replacement. Starting with an IoU of 86.47\% at 0\% missing data, SMAGNet maintained a highest score of 79.53\% even when 100\% of the MSI data was replaced with missing data. With a smallest performance degradation of 6.94\%, SMAGNet demonstrates superior capability in handling scenarios where MSI data are partially or entirely missing.

\begin{table}[t]
\centering
\caption{Experimental result of the robustness study.}
\label{tab3}
\resizebox{\textwidth}{!}{%
\begin{tabular}{lccccccl}
\toprule
\multirow{2}{*}{Model} &
  \multicolumn{5}{c}{IoU with varying missing data pixel replacement ratios in MSI data (\%) \vspace{6ex}} &
  \multirow{2}{*}{\begin{tabular}[c]{@{}c@{}} $\Delta$ \\ (0\% -\\ 100\%)\end{tabular}} &
  \multicolumn{1}{c}{\multirow{2}{*}{\begin{tabular}[c]{@{}c@{}}P-value \\ (100\% missing \\ data in MSI data \\ vs. U-Net (SAR))\end{tabular}}} \\ \cline{2-6}
           & 0\%           & 25\%          & 50\%          & 75\%           & 100\%          &       & \multicolumn{1}{c}{} \\ \hline\hline
VFuseNet   & 83.33 (±1.00) & 79.02 (±1.57) & 74.60 (±2.57) & 70.22 (±3.44)  & 65.97 (±3.98)  & 17.36 & \qquad 0.000 ***            \\ \hline
FuseNet    & 83.40 (±1.13) & 79.49 (±1.98) & 75.36 (±2.82) & 71.23 (±3.58)  & 67.24 (±4.10)  & 16.16 & \qquad 0.000 ***            \\ \hline
DeepLabV3+ & 84.48 (±1.19) & 82.43 (±1.81) & 79.22 (±2.69) & 74.19 (±3.99)  & 67.33 (±5.75)  & 17.15 & \qquad 0.000 ***            \\ \hline
FTransUNet & 83.93 (±2.64) & 78.10 (±5.47) & 75.01 (±8.42) & 72.26 (±10.87) & 67.50 (±11.70) & 16.43 & \qquad 0.000 ***            \\ \hline
U-Net++    & 84.41 (±1.54) & 80.65 (±2.75) & 76.44 (±4.08) & 71.99 (±5.67)  & 67.77 (±6.86)  & 16.63 & \qquad 0.000 ***            \\ \hline
PSPNet     & 82.65 (±0.85) & 81.16 (±2.62) & 78.10 (±5.55) & 73.60 (±8.32)  & 68.40 (±9.60)  & 14.25 & \qquad 0.000 ***            \\ \hline
U-Net      & 84.96 (±0.97) & 81.62 (±1.56) & 77.83 (±2.09) & 73.89 (±2.66)  & 69.97 (±3.11)  & 14.99 & \qquad 0.000 ***            \\ \hline
FPN        & 84.25 (±0.96) & 82.74 (±1.61) & 80.16 (±2.02) & 76.64 (±2.76)  & 70.54 (±6.08)  & 13.71 & \qquad 0.000 ***            \\ \hline
MFGFUnet   & 85.96 (±0.57) & 83.88 (±0.84) & 81.11 (±1.28) & 77.80 (±1.89)  & 72.98 (±3.05)  & 12.97 & \qquad 0.000 ***            \\ \hline
MCANet     & 85.48 (±0.99) & 83.87 (±1.05) & 81.86 (±1.41) & 78.70 (±2.16)  & 74.71 (±3.25)  & 10.77 & \qquad 0.001 ***            \\ \hline
CMGFNet    & 84.70 (±0.59) & 82.37 (±0.89) & 80.32 (±1.19) & 78.17 (±1.52)  & 76.34 (±1.80)  & 8.36  & \qquad 0.000 ***            \\ \hline
CMFNet     & 84.95 (±0.87) & 84.20 (±0.65) & 82.64 (±1.12) & 80.43 (±1.50)  & 77.92 (±1.21)  & 7.03  & \qquad 0.002 **             \\ \hline
SMAGNet (Ours) &
  \textbf{86.47} (±0.61) &
  \textbf{84.70} (±0.80) &
  \textbf{83.07} (±0.99) &
  \textbf{81.17} (±1.16) &
  \textbf{79.53} (±1.28) &
  \textbf{6.94} &
  \qquad 0.850 \\ \bottomrule
\end{tabular}%
}
\end{table}

Following SMAGNet, CMFNet, CMGFNet, MFGUNet, and MCANet also achieved high IoU scores in the robustness study, though their IoU rankings varied across scenarios depending on the proportion of missing data pixels in MSI data. Specifically, CMFNet showed the second-highest robustness when the missing data ratio was 25\% or higher. MFGUNet and MCANet also presented strong performance with original MSI data but were less robust than SMAGNet, with performance drops of 12.97\% and 10.77\%, respectively. In terms of the standard deviation of IoU, all multimodal deep learning models exhibited progressively larger variability as the proportion of missing MSI data pixels increased. SMAGNet maintained a relatively stable IoU standard deviation compared to other models, even as it followed the same pattern of increasing variability.

\begin{table}[t]
\centering
\caption{Performance comparison between SMAGNet (SAR-only), which refers to SMAGNet under the condition of 100\% missing MSI data, and U-Net (SAR).}
\label{tab4}
\resizebox{\columnwidth}{!}{%
\begin{tabular}{lcccc}
\toprule
\multicolumn{1}{c}{Model}              & IoU (\%)      & Precision (\%) & Recall (\%)   & OA (\%)       \\ \hline\hline
U-Net (SAR)        & 79.65 (±0.96) & 90.81 (±0.83)  & 86.64 (±1.03) & 96.52 (±0.18) \\ \hline
SMAGNet (SAR-only) & 79.53 (±1.28) & 91.36 (±1.30)  & 86.02 (±1.62) & 96.52 (±0.23) \\ \bottomrule
\end{tabular}%
}
\end{table}

In Table \ref{tab3}, we reported the p-values obtained from the Mann-Whitney U test \citep{mann1947test} to examine the statistical significance of the difference in IoU between the U-Net (SAR) and the multimodal deep learning models in cases where MSI data was replaced with 100\% missing data. This case represents situations where only SAR data can be leveraged for post-flood water mapping at inference time. The analysis results showed statistically significant differences in most models’ prediction results (p-value $<$ 0.05), indicating that when MSI data is 100\% missing, the performance of other multimodal deep learning models is lower than that of the U-Net trained on SAR alone. This means that other comparative multimodal models do not effectively account for edge cases where supplementary modalities may not be available. A notable exception was the SMAGNet model, which shows statistically comparable results with the U-Net (SAR) model in this scenario. These results suggest that deploying multimodal deep learning models in real-world post-flood water mapping scenarios without addressing pixel-level missing data may lead to significant performance degradation compared to single-modality approaches, whereas SMAGNet demonstrates strong effectiveness even under such challenging conditions. To support these findings, Table \ref{tab4} presents the four performance metrics of U-Net (SAR) and SMAGNet under the condition where MSI data is entirely missing. The experimental results presented in Table \ref{tab3} and Table \ref{tab4} show that, despite being designed to utilize both SAR and MSI data, SMAGNet achieves comparable performance to the U-Net (SAR) model when using only SAR data.

The statistical test results are also closely aligned with the training and validation loss curves for SMAGNet and U-Net (SAR) illustrated in Fig. \ref{fig7}. The training and validation losses for the segmentation head using SAR-only features in SMAGNet converge to loss values comparable to those of U-Net (SAR). Furthermore, the training and validation losses for the segmentation head using SAR-MSI fused features in SMAGNet are lower than the losses observed in both the U-Net (SAR) model and the SMAGNet model with SAR-only features. These observations demonstrate that SMAGNet was trained to a comparable performance level as U-Net (SAR) on SAR features and simultaneously trained to achieve superior performance on SAR-MSI fused features compared to U-Net (SAR).

\subsection{Ablation Study}
\label{Section5.3}

We conducted an ablation study to assess the contribution of two key components in SMAGNet to the performance improvement: (1) the spatial mask in SMAG-FFM and (2) the weight-shared decoder. Table \ref{tab5} presents the results of the ablation study for SMAGNet on IoU scores under varying levels of incomplete MSI data, following the same settings as in Section \ref{Section5.2}. In the ablation study, Case (a) represents SMAGNet without the spatial mask and the weight-shared decoder. This model is equivalent to one that employs the gating mechanism for feature fusion, as described in \cite{hosseinpour2022cmgfnet}, and includes two independent decoders for SAR-only features and SAR-MSI fused features. With the original MSI data, Case (a) achieved an IoU of 85.77\%. However, when all MSI data pixels are replaced with missing data, performance drops to 75.86\%, resulting in a degradation of 9.91\%.

\begin{table}[b]
\centering
\caption{Experimental result of the ablation study.}
\label{tab5}
\resizebox{\textwidth}{!}{%
\begin{tabular}{ccccccccc}
\toprule
\multirow{2}{*}{Case} &
  \multirow{2}{*}{\begin{tabular}[c]{@{}c@{}}Weight-shared\\ Decoder\end{tabular}} &
  \multirow{2}{*}{\begin{tabular}[c]{@{}c@{}}Spatial\\ Mask\end{tabular}} &
  \multicolumn{5}{c}{IoU with varying missing data pixel replacement ratios in MSI data (\%)} &
  \multirow{2}{*}{\begin{tabular}[c]{@{}c@{}} $\Delta$ \\ (0\% - 100\%)\end{tabular}} \\ \cline{4-8}
 &
   &
   &
  0\% &
  25\% &
  50\% &
  75\% &
  100\% &
   \\ \hline\hline
(a) &
  \multicolumn{1}{l}{} &
  \multicolumn{1}{l}{} &
  85.77 (±0.68) &
  83.29 (±0.84) &
  80.85 (±1.28) &
  78.10 (±1.85) &
  75.86 (±2.27) &
  9.91 \\ \hline
(b) &
  \checkmark &
  \multicolumn{1}{l}{} &
  85.61 (±0.86) &
  83.58 (±0.86) &
  81.52 (±0.95) &
  79.30 (±1.22) &
  77.11 (±1.42) &
  8.50 \\ \hline
(c) &
  \checkmark &
  \checkmark &
  \textbf{86.47} (±0.61) &
  \textbf{84.70} (±0.80) &
  \textbf{83.07} (±0.99) &
  \textbf{81.17} (±1.16) &
  \textbf{79.53} (±1.28) &
  6.94 \\ \bottomrule
\end{tabular}%
}
\end{table}

In Case (b), the replacement of the two independent decoders with the weight-shared decoder enhances the model's robustness in handling missing data pixels. The IoU with the original MSI data remained similar at 85.61\%, but when all MSI data pixels were missing, the IoU increased from 75.86\% to 77.11\%. This reduced the performance drop from 9.91\% to 8.5\%. This result demonstrates that the weight-shared decoder contributes to mitigating the performance degradation caused by missing data pixels in MSI data. 

Furthermore, in Case (c), SMAGNet, which integrates both the weight-shared decoder and the spatial mask, achieved the best performance compared to the models in Case (a) and (b) across all scenarios with varying levels of missing data. For instance, Case (c) reached the highest IoU of 86.47\% with the original MSI data and 79.53\% when all MSI data pixels were missing. This configuration also showed the smallest performance degradation at 6.94\%. These results highlight the performance improvements achieved through the proposed strategies and demonstrate the model's robustness in handling missing data.

\begin{figure}[p]
  \centering
  \includegraphics[width=0.75 \linewidth]{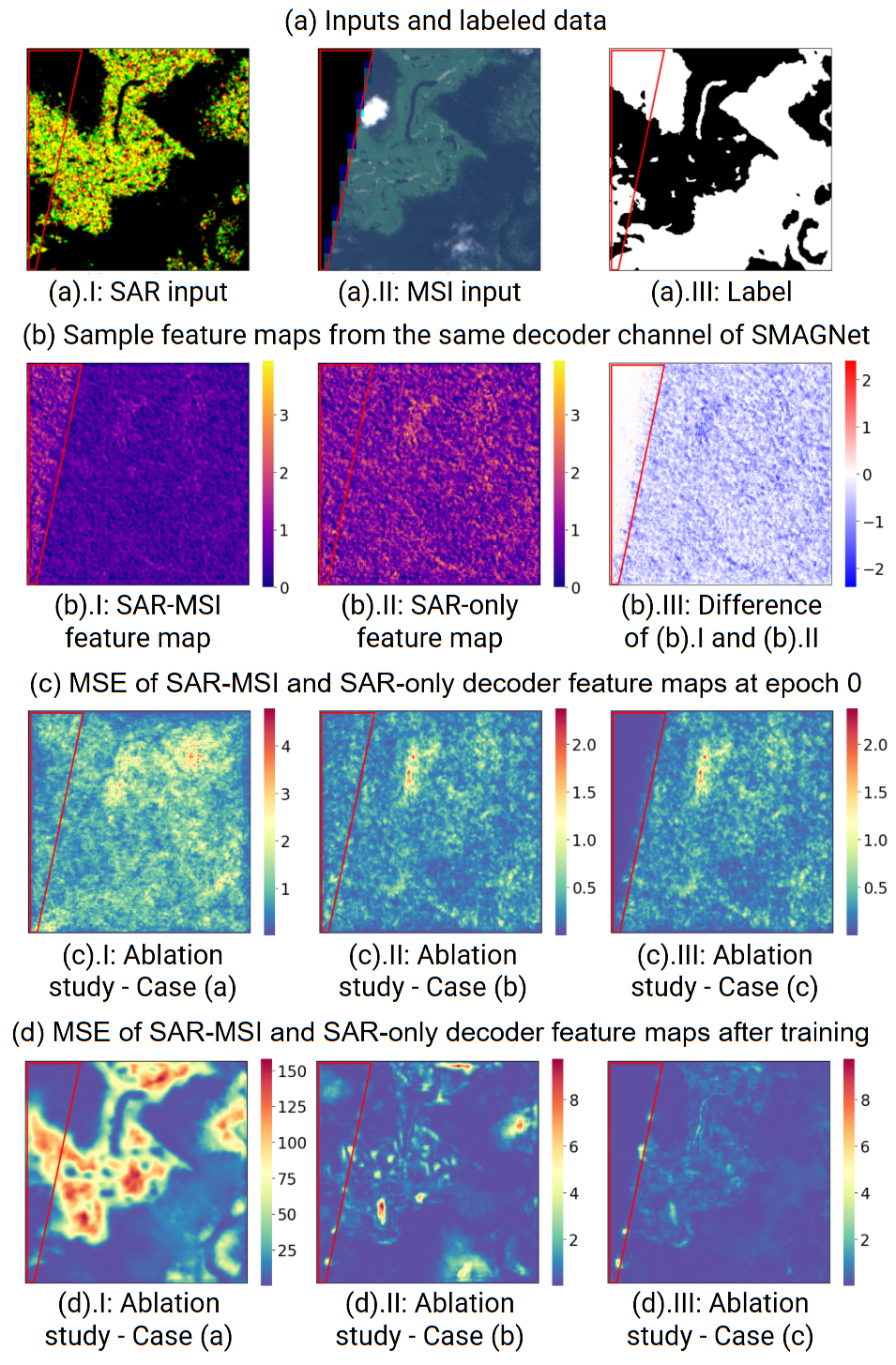}
  \caption{Mean squared error (MSE) between the SAR-MSI fused and SAR-only decoder output feature maps at the initial epoch and after training. The region outlined by the red solid line indicates missing data in the MSI data. (a) Inputs and labeled data: (a).I SAR input, (a).II MSI input with missing pixels (black), (a).III labeled data (flooded area: white, background: black). (b) Decoder output feature maps at the initial epoch for Case (c): (b).I SAR–MSI path, (b).II SAR-only path, (b).III feature map difference. (c) MSE between decoder outputs at the initial epoch in the ablation study: (c).I Case (a), (c).II Case (b), (c).III Case (c). (d) MSE between decoder outputs after training in the ablation study: (d).I Case (a), (d).II Case (b), (d).III Case (c).}
  \label{fig10}
\end{figure}

To provide deeper insights into how the combination of the two components effectively addresses missing data pixels in MSI data, we present visualizations based on the output feature maps from the decoder and the gate maps in SMAGNet. Fig. \ref{fig10} illustrates the mean squared error (MSE) between the SAR-MSI fused output feature map and the SAR-only output feature map from the decoder in SMAGNet, both at the initial epoch and after training, using sample input data (Fig. \ref{fig10} (a)). In particular, Fig. \ref{fig10} (b) presents an illustrative visualization showing that, in SMAGNet, the feature vectors corresponding to the missing data regions of the MSI data in the SAR-MSI fused output feature map (Fig. \ref{fig10} (b).I) are very similar to those covering the same regions of the SAR-only output feature map (Fig. \ref{fig10} (b).II), as indicated by the near-zero difference in the solid red line region (Fig. \ref{fig10} (b).III).

The results of the differences between the two output feature maps are presented in Fig. \ref{fig10} (c) and (d). Fig. \ref{fig10} (c) shows the visualization at the initial epoch, while Fig. \ref{fig10} (d) displays the visualization after the training is complete. Both figures present the MSE results between the two output feature maps for the three cases used in the ablation study. In Case (a), when the spatial mask and a weight-shared decoder were not applied, as shown in Fig. \ref{fig10} (c).I and (d).I, the differences between the two output feature maps are larger than those in Case (b) and Case (c). Therefore, Fig. \ref{fig10} (c).I and (d).I particularly indicate that, despite the missing data pixels in the MSI data being irrelevant for post-flood water mapping, the features extracted from these missing data pixels influence the prediction results in the configuration with two separate decoders.

On the other hand, as shown in Fig. \ref{fig10} (c).II and (d).II, when calculating the MSE between the two output feature maps in Case (b), we observed a decreased difference than Fig. \ref{fig10} (c).I and (d).I. Furthermore, as shown in Fig. \ref{fig10} (c).III and (d).III, the feature vectors corresponding to the areas with missing data pixels in the MSI data exhibited almost no difference, both at the initial epoch and after training. Consequently, Fig. \ref{fig10} (c).III and (d).III strongly support that the spatial mask effectively filters out features extracted from the missing data pixels in MSI data and preserves SAR features. These visualizations also illustrate that both SAR-only features and SAR-MSI fused features are integrated and processed through a weight-shared decoder. 

\begin{figure}[t]
  \centering
  \includegraphics[width=1.0 \linewidth]{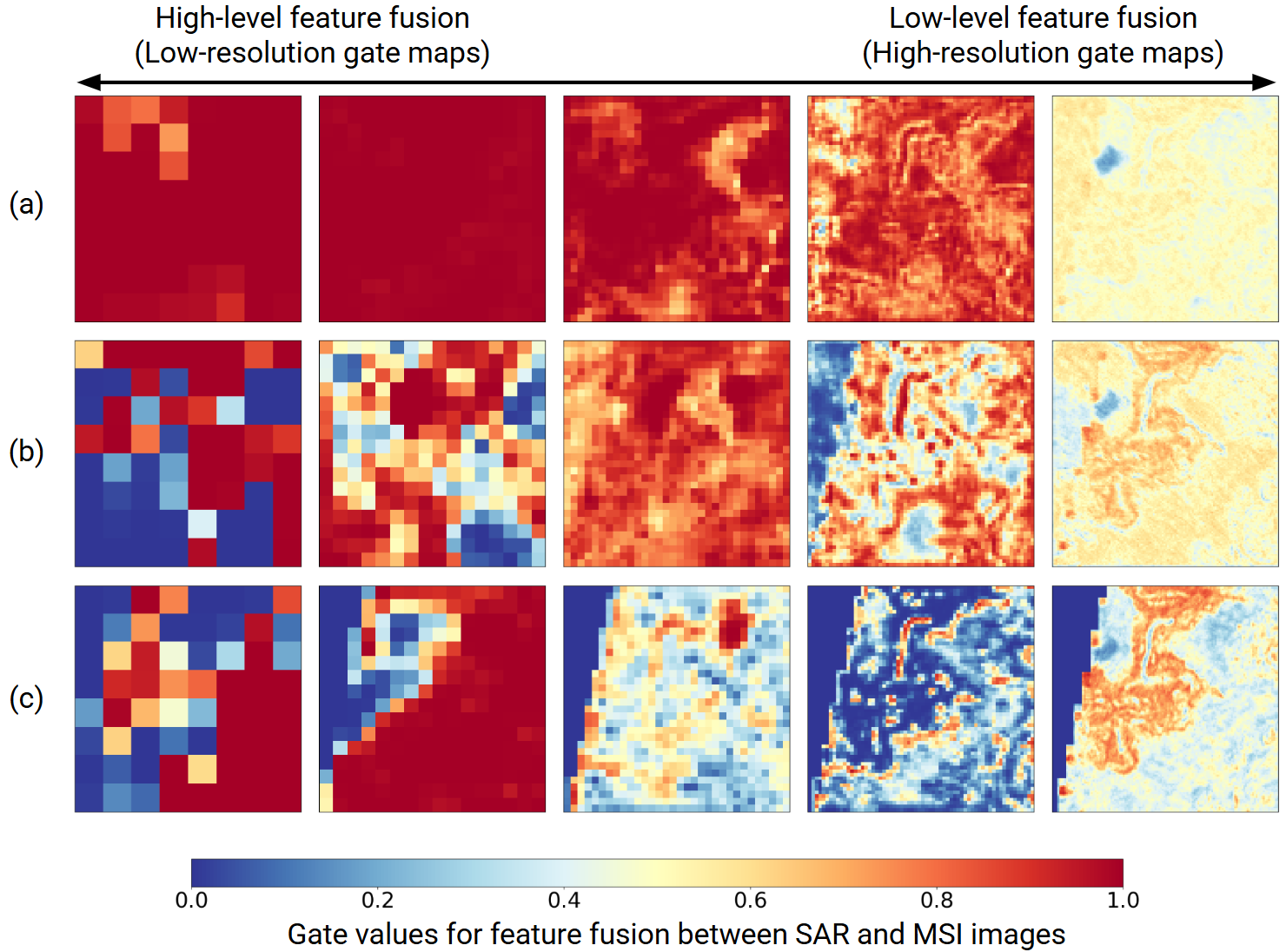}
  \caption{Visualization of gate maps at five different feature scales. Cases (a), (b), and (c) correspond to the three cases presented in Table \ref{tab5}.}
  \label{fig11}
\end{figure}

Fig. \ref{fig11} illustrates the visualization of gate maps at five different feature scales for the three cases in the ablation study. The columns represent different levels of feature scale, ranging from high-level (low-resolution gate maps) on the left to low-level (high-resolution gate maps) on the right. The rows (a), (b), and (c) correspond to the three cases in the ablation study. The color scale at the bottom indicates activation of gate maps for MSI features in feature fusion, with blue representing lower gate values (close to 0) and red representing higher gate values (close to 1). As the resolution of the gate maps increases from left to right, they can adjust the contributions of SAR and MSI features at finer levels of detail. In Case (c), where both the spatial mask and a weight-shared decoder are applied together, the gate activations exhibit more distinct contrast, particularly in high-resolution gate maps, compared to Case (a) and (b). This pronounced contrast suggests that the gate map has been effectively trained to allocate distinct contributions of SAR and MSI features in the fusion process.

\subsection{Generalizability Study}
\label{Section5.4}

Generalizability studies in practical scenarios are essential for evaluating a model's applicability to real-world conditions. SMAGNet was specifically designed to address the practical challenge of partially available MSI data in SAR-based post-flood water mapping. Therefore, this section evaluates the generalizability of SMAGNet using a real-world flood event not seen during training.

\begin{figure}[!b]
  \centering
  \includegraphics[width=1.0 \linewidth]{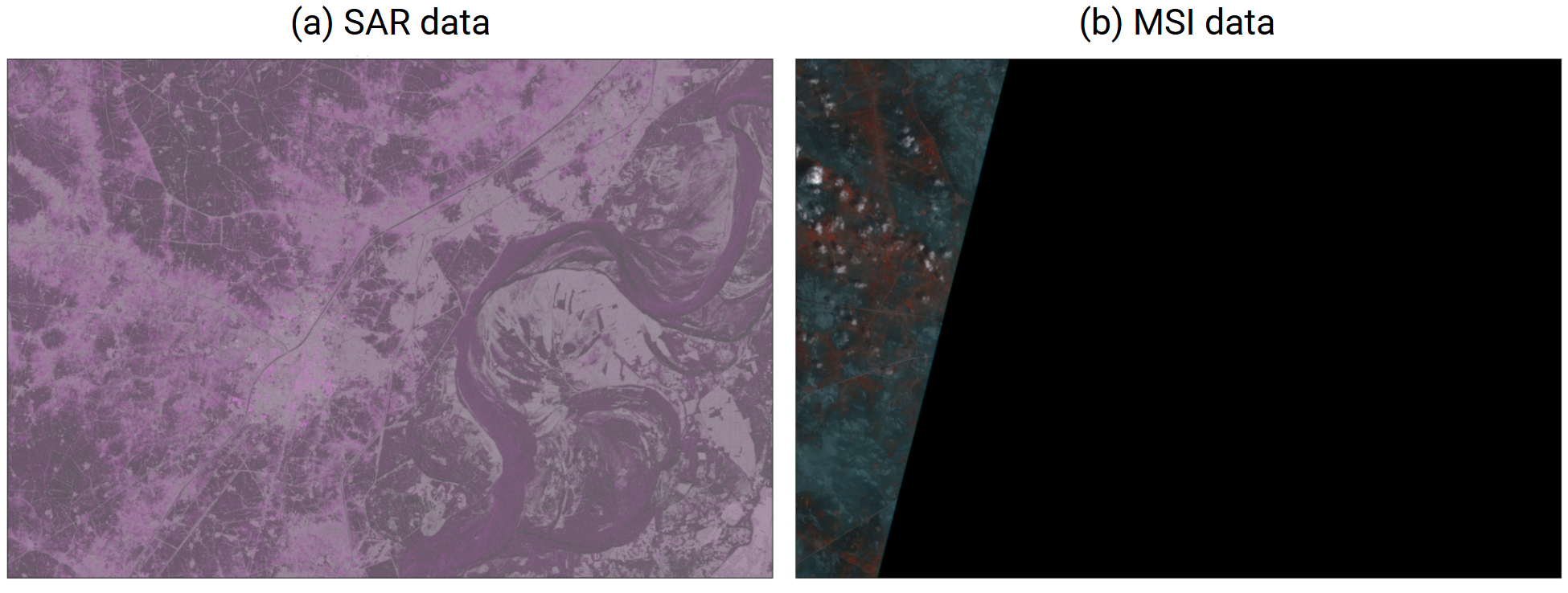}
  \caption{SAR data (August 30, 2022) and MSI data (August 29, 2022) over Larkana, Pakistan. The MSI data, visualized in a false-color composite using NIR, Green, and Red channels, contains 20\% valid pixels and 80\% missing pixels.}
  \label{fig12}
\end{figure}

To construct a dataset for the generalizability study that does not coincide spatially and temporally with the training data, we used the STURM-Flood dataset, which contains SAR data and corresponding labels for post-flood water mapping \citep{notarangelo2025sturm}. We excluded flood events from the STURM-Flood dataset that occurred between 2016 and 2020, as this period overlaps with the temporal coverage of the training dataset. We then selected SAR data from the STURM-Flood dataset that had corresponding MSI observations available one day prior, along with the corresponding labels. As a result, SAR data collected over Larkana, Pakistan, on August 30, 2022, along with the corresponding labels, were used as the dataset for the generalizability study. Specifically, Sentinel-2 MSI data acquired one day prior over the same region were obtained from Google Earth Engine. The MSI data contained valid pixels for approximately 20\% of the area, with the remaining 80\% comprising missing values (see Fig. \ref{fig12}). The SAR and MSI data, each with a size of 3,584 × 2,432 pixels, were divided into 256 × 256 tiles, resulting in 126 tiles for the generalizability study. This experiment was conducted using the same trained models as those used in the comparative study in Section \ref{Section5.1}, with the only difference being the use of the generalizability study dataset.

\begin{table}[t]
\caption{Experimental result of the generalizability study.}
\label{tab6}
\resizebox{\columnwidth}{!}{%
\begin{tabular}{lcccc}
\toprule
\multicolumn{1}{c}{Model}          & IoU (\%)               & Precision (\%)         & Recall (\%)            & OA (\%)                \\ \hline\hline
U-Net (SAR)    & 61.78 (±5.03)          & \textbf{92.08} (±0.63) & 65.30 (±5.72)          & 75.78 (±3.03)          \\ \hline
FTransUNet     & 28.76 (±9.33)          & 90.38 (±4.40)          & 30.08 (±10.71)         & 55.79 (±4.81)          \\ \hline
U-Net++        & 56.38 (±7.36)          & 84.88 (±2.81)          & 62.95 (±9.31)          & 70.91 (±4.43)          \\ \hline
DeepLabV3+     & 56.93 (±9.60)          & 84.69 (±5.29)          & 65.03 (±14.96)         & 71.16 (±4.78)          \\ \hline
U-Net          & 58.83 (±4.07)          & 84.99 (±2.61)          & 65.93 (±6.27)          & 72.40 (±2.19)          \\ \hline
MCANet         & 56.63 (±8.71)          & 89.77 (±3.64)          & 60.33 (±9.05)          & 72.19 (±5.86)          \\ \hline
CMFNet         & 63.21 (±4.05)          & 84.57 (±3.39)          & \textbf{71.75} (±6.29) & 75.00 (±2.49)          \\ \hline
PSPNet         & 62.23 (±6.40)          & 85.65 (±3.63)          & 70.07 (±9.34)          & 74.68 (±3.51)          \\ \hline
FPN            & 60.79 (±9.81)          & 88.88 (±3.53)          & 66.61 (±13.25)         & 74.58 (±5.52)          \\ \hline
CMGFNet        & 58.10 (±7.95)          & 90.27 (±1.97)          & 62.26 (±9.71)          & 73.20 (±4.65)          \\ \hline
MFGFUNet       & 61.83 (±4.43)          & 84.68 (±2.72)          & 69.82 (±6.15)          & 74.19 (±2.74)          \\ \hline
FuseNet        & 63.78 (±4.88)          & 85.75 (±1.75)          & \underline{71.42} (±6.15)    & 75.69 (±3.10)          \\ \hline
VFuseNet       & \underline{64.14} (±7.17)    & 87.00 (±2.75)          & 71.19 (±9.18)          & \underline{76.21} (±4.42)    \\ \hline
SMAGNet (Ours) & \textbf{64.70} (±6.24) & \underline{90.78} (±1.95)    & 69.39 (±7.66)          & \textbf{77.33} (±3.84) \\ \bottomrule
\end{tabular}%
}
\end{table}

As presented in Table \ref{tab6}, SMAGNet achieved the highest IoU (64.70\%) and overall accuracy (77.33\%), demonstrating the strong generalizability of SMAGNet in real-world flood events with partially available MSI data. Although U-Net (SAR) achieved the highest precision (92.08\%), it showed relatively lower recall (65.30\%) and IoU (61.78\%), indicating a tendency to produce fewer false positives but more false negatives in post-flood water areas. By contrast, FuseNet and CMFNet achieved higher recall values (71.42\% and 71.75\%, respectively), but their overall accuracy and IoU were lower than those of SMAGNet. These results highlight the effectiveness of SMAGNet in utilizing partially available MSI data to complement SAR observations, thereby enabling more accurate delineation of post-flood water extent under practical conditions.

\section{Discussion}
\label{Section6}

Through the experiments in Section \ref{Section5.1}, we demonstrated the superior performance of SMAGNet in terms of IoU, Recall, and Accuracy, achieving highest scores of 86.47\% for IoU, 92.45\% for Recall, and 97.73\% for Accuracy. In addition, we showed that the incorporation of MSI data reduced the number of misclassified pixels across the majority of NDVI ranges in SAR-based post-flood water mapping. However, in densely vegetated areas, the difference in false negatives between SMAGNet and U-Net (SAR) was negligible, with an IoU score difference of less than 0.01\%. This indicates that the additional spectral information from MSI may be less effective in distinguishing post-flood water in areas with dense vegetation. These findings suggest the necessity for research into alternative sensors that are effective for post-flood water detection in densely vegetated areas.

Moreover, in Section \ref{Section5.2}, SMAGNet consistently exhibited robust performance in handling incomplete MSI data under various conditions, where 25\% to 100\% of the pixels were replaced with missing data. Notably, our statistical tests showed that SMAGNet performed comparably to the U-Net (SAR) with no significant difference, even when using MSI data with 100\% missing data pixels. This result suggests that SMAGNet effectively leverages MSI data under varying availability conditions, while maintaining robustness. This robustness in handling partially available MSI data demonstrates the practical applicability of SMAGNet, indicating that the advantages of multimodal deep learning can be utilized even when MSI data is incomplete at inference time.

The ablation study provides strong evidence that the combination of the spatial mask in SMAG-FFM and the weight-shared decoder is effective in addressing the missing data present in MSI data for post-flood water mapping. Specifically, the visualization results presented in Fig. \ref{fig10} illustrated that the spatial mask filters out feature vectors extracted from MSI data in regions where missing data pixels are present, while preserving feature vectors extracted from SAR data during the feature fusion process. In other words, even when feature fusion occurs between SAR and MSI data, the SAR feature vectors are preserved and forwarded to the weight-shared decoder in regions where missing data pixels are present in the MSI data. As a result, the two output feature maps from the decoder in SMAGNet yield identical SAR-only feature vectors in areas where missing pixels are present in the MSI data, thereby enhancing robustness to missing data in the final output by minimizing the impact of feature vectors extracted from the missing data pixels in MSI data.

Furthermore, the weight-shared decoder contributes to robust predictive performance for missing data pixels in the MSI data by simultaneously learning both SAR-only and SAR-MSI fused feature representations. The training and validation loss graphs shown in Fig. \ref{fig7} illustrate that the weighted-shared decoder in SMAGNet effectively captured these two types of representations. Specifically, in both training and validation, the loss for predictions using SAR-only features is similar to that of the U-Net trained solely on SAR data. In contrast, predictions made with SAR-MSI fused features in SMAGNet showed lower loss values than the previous two loss values. This suggests that the weight-shared decoder enables SMAGNet to effectively leverage both SAR-only and SAR-MSI fused features, achieving robust performance despite incomplete MSI data.

In SMAGNet, the Spatially Masked Gate (SMG) map is a core component that adaptively fuses SAR and MSI features while filtering out missing data, which is essential for enhancing performance. In the ablation study, it was observed that among the three combinations, the SMG maps of SMAGNet showed the highest overall activation levels, with each region for the SAR and MSI features contributing prominently and exhibiting a strong contrast, as shown in Fig. \ref{fig11} (c). On the contrary, in the other two cases shown in Fig. \ref{fig11} (a) and \ref{fig11} (b), features from regions with missing data directly influenced the prediction results, and we observed that the highest resolution SMG map was trained with relatively similar contributions between SAR and MSI features across regions. This similar level of contribution implies that the unique characteristics of SAR and MSI features were not efficiently utilized in a complementary manner for prediction.

In the generalizability study, experimental results highlight the effectiveness of SMAGNet in enhancing SAR-based post-flood water mapping using incomplete MSI data in a real-world scenario. While SMAGNet achieved the highest performance in terms of IoU and overall accuracy in the generalizability study, the overall performance of all models decreased compared to the results obtained in the comparative study. This degradation is potentially caused by domain shift, highlighting the importance of spatially and temporally diverse training data to improve model generalizability.

A limitation of the current approach is that SMAGNet primarily focuses on post-flood water mapping rather than fine-grained flood damage segmentation (e.g., distinguishing flooded roads, buildings, or agricultural fields). Although combining post-flood extent maps with pre-flood data (e.g., Land Use and Land Cover (LULC), road networks, and building footprints) can assist in estimating damage, future research would be valuable in developing dedicated deep learning models for detailed, class-specific segmentation of flooded land cover types to improve flood damage assessment and support recovery planning. In addition, this study is constrained by the lack of diverse benchmark datasets for multimodal deep learning in post-flood water mapping. This scarcity of benchmark datasets hinders a more robust evaluation of the model performance across different benchmark datasets. To compensate for this limitation, we performed 10 repeated experiments to report reliable performance assessment on the C2S-MS Flood dataset.

\section{Conclusion}
\label{Section7}

In the flood management cycle, especially during the response stage, the provision of timely and accurate information is essential. SAR-based post-flood water mapping has the advantage of being able to observe the Earth's surface even during cloud-covered flood events, enabling the mapping of floodwater extent. By integrating SAR data with available MSI data, multimodal deep learning models can further enhance the accuracy of post-flood water mapping. However, these models are required to be robust against missing data pixels in MSI data, which frequently occur in practical scenarios. To address this research gap, we proposed the Spatially Masked Adaptive Gated Network (SMAGNet). In our experiments with the C2S-MS Floods dataset, SMAGNet consistently outperforms other multimodal deep learning models on prediction performance in various scenarios where different proportions of MSI data pixels are replaced with missing data. Furthermore, we found that even when all MSI data were missing, the performance of SMAGNet remained comparable to that of a U-Net trained solely on SAR data without statistically significant difference. These findings indicate that SMAGNet enhances the robustness to missing data as well as the applicability of multimodal deep learning in real-world flood management scenarios. For future research, extending SMAGNet to fine-grained flood damage segmentation tasks, such as distinguishing between flooded roads, buildings, and agricultural fields, could enhance the effectiveness of damage assessment and recovery planning by leveraging the increased spatial and temporal resolution of satellite imagery and the advanced capabilities of deep learning.





\bibliographystyle{elsarticle-harv} 
\bibliography{2024-SMAGNet-reference}

\newpage

\listoffigures





\end{document}